%% file: main.tex
\documentclass[twoside]{article}

\usepackage[accepted]{aistats2022}
\usepackage[round]{natbib}

\usepackage{csquotes}
\usepackage{epsfig}
\usepackage[dvipsnames]{xcolor}

\usepackage{balance} 
\usepackage{algorithmic}
\usepackage{algorithm}
\usepackage{multirow}
\usepackage{graphicx}
\usepackage{amsmath}
\usepackage{amssymb}
\usepackage{amsfonts}
\usepackage{xspace}
\usepackage{url}
\usepackage{booktabs}
\usepackage{bbold}
\usepackage{tcolorbox, changepage, dsfont}
\usepackage{pifont}
\usepackage[dvipsnames]{xcolor}

\input{math_commands}

\usepackage{lipsum,graphicx,subcaption}

\newcommand{\IGNORE}[1]{}

\newcommand{\normaltilde}{\raise.17ex\hbox{$\scriptstyle\sim$}}

\begin{document}

%
\runningtitle{Fast Sparse Classification for Generalized Linear and  Additive Models}

%

\twocolumn[

\aistatstitle{Fast Sparse Classification for Generalized Linear and \\ Additive Models}

\aistatsauthor{ Jiachang Liu$^{1}$ \And Chudi Zhong$^{1}$ \And  Margo Seltzer$^{2}$ \And Cynthia Rudin$^{1}$  }

\aistatsaddress{ $^{1}$Duke University \quad  $^{2}$ University of British Columbia \\
\texttt{\{jiachang.liu, chudi.zhong\}@duke.edu, mseltzer@cs.ubc.ca, cynthia@cs.duke.edu}} ]

\input{sections/00-abstract}
\input{sections/01-introduction}
\input{sections/02-background}

\input{sections/03-overview}
\input{sections/04-cuts}
\input{sections/05-exponential}
\input{sections/06-dynamic}

\input{sections/07-experiment}

\input{sections/08-relatedWorks}
\input{sections/09-conclusion}

\bibliography{bibliography}
\bibliographystyle{plainnat}


\clearpage
\appendix

\thispagestyle{empty}

\onecolumn \makesupplementtitle

\input{sections/10-supplement}

\end{document}

%% file: math_commands.tex
\usepackage{amsmath}
\usepackage{amsthm}
\usepackage{amsfonts}
\usepackage{amssymb}
\usepackage{bm}
\usepackage{graphicx}
\usepackage{sidecap}
\usepackage{mathrsfs}
\usepackage{multirow, multicol}
\usepackage{caption}

\usepackage{wrapfig}
\usepackage{booktabs}

\usepackage{natbib} 
\newtheorem{theorem}{Theorem}[section]

\newcommand{\bc}{\bm{c}}

\newcommand{\bw}{\bm{w}}
\newcommand{\bx}{\bm{x}}
\newcommand{\bz}{\bm{z}}

\newcommand{\be}{\bm{e}}

\def\1{\bm{1}}

\DeclareMathAlphabet{\mathsfit}{\encodingdefault}{\sfdefault}{m}{sl}
\SetMathAlphabet{\mathsfit}{bold}{\encodingdefault}{\sfdefault}{bx}{n}

\DeclareMathOperator*{\argmin}{arg\,min}

%% file: sections/00-abstract.tex
\begin{abstract}
  We present fast classification techniques for sparse generalized linear and additive models. These techniques can handle thousands of features and thousands of observations in minutes, even in the presence of many highly correlated features. For fast sparse logistic regression, our computational speed-up over other best-subset search techniques owes to linear and quadratic surrogate cuts for the logistic loss that allow us to efficiently screen features for elimination, as well as use of a priority queue that favors a more uniform exploration of features. As an alternative to the logistic loss, we propose the exponential loss, which permits an analytical solution to the line search at each iteration.
  Our algorithms are generally 2 to 5 times faster than previous approaches. They produce interpretable models that have accuracy comparable to black box models on challenging datasets.
\end{abstract}

%% file: sections/01-introduction.tex
\section{INTRODUCTION}
Our goal is to produce sparse generalized linear models or sparse generalized additive models from large datasets in under a minute, even in the presence of highly-correlated features. Specifically, our interest is in the following 
problem:
\begin{equation}
\label{l0_loss}
    \begin{aligned}
    	\min_{\bw} &\sum_{i=1}^{n} \ell(\bw, \bx_i, y_i) + \lambda_{0} \lVert \bw \rVert_{0} 
    \end{aligned}
\end{equation}
with the logistic loss $$\ell(\bw, \bx_i, y_i)=\log\left(1+e^{-y_{i}(\bw^{T}\bx_{i})}\right)$$
or the exponential loss
$$\ell(\bw, \bx_i, y_i)=e^{-y_{i}(\bw^{T}\bx_{i})}$$
where $\bx_{i} \in \mathbb{R}^{p}$ is the $i$-th observation, and $y_{i} \in \{-1, 1\}$ is the label of the $i$-th data sample. The logistic loss tends to yield nicely calibrated probability estimates, which explains its broad appeal. The exponential loss, used in boosting, has been overlooked as an approach to sparse additive modeling, but like logistic regression, it also yields direct probability estimates. It has the advantage of analytical solutions for line search, dramatically improving convergence rates.

A small $\ell_2$ regularization is used with the logistic loss to speed up convergence, as discussed later: 
\begin{equation}
\label{l0_l2_loss}
    \begin{aligned}
    	\min_{\bw} &\sum_{i=1}^{n} \ell(\bw, \bx_i, y_i) + \lambda_{0} \lVert \bw \rVert_{0} + \lambda_{2} \lVert \bw \rVert_{2}^2.
    \end{aligned}
\end{equation}
We do not include $\ell_1$: since we are looking for very sparse and accurate models, $\ell_1$ regularization would degrade the quality of the solution compared to true sparsity regularization with $\ell_0$. The $\ell_{0}$ penalty term makes Problems~\eqref{l0_loss} or \eqref{l0_l2_loss} NP-hard. 

Problems~\eqref{l0_loss} or \eqref{l0_l2_loss} can produce generalized additive models~\citep{lou2016sparse, hastie2017generalized, nori2019interpretml, rudin2022interpretable} through a transformation of the input variables, replacing each continuous feature $x_{\cdot,j}$ with a set of dummy variables $\tilde{x}_{\cdot,j,\theta} = \bm{1}_{[x_{\cdot,j}\geq \theta]}$, for $\theta$ set to be each realized value of feature $j$ in the dataset. Then, solving \eqref{l0_loss} or \eqref{l0_l2_loss} yields a generalized additive model where component function $j$ is a sum of the weighted dummy variables for feature $j$. 
This transformation yields a large feature set with many correlated features, but the approaches provided here can handle such sizes. 


There are at least two general approaches for tackling these problem (besides relaxing the $\ell_0$ term to $\ell_1$ and suffering the associated bias).
The first uses callbacks to a mathematical programming solver, such as a mixed-integer programming (MIP) solver~\citep{sato2016feature, ustun2017optimized, sato2017piecewise, bertsimas2017logistic, bertsimas2017sparse, ustun2019learning}.
This approach can solve the problem exactly.
However, it cannot handle large feature spaces or highly-correlated features. 
A solver might take several days or run out of memory on even a modestly-sized problem.
The second approach to Problems~\eqref{l0_loss} or \eqref{l0_l2_loss} is to use coordinate descent with local swap operations for best subset search, similar to simulated annealing, Metropolis-Hastings, or other MCMC methods~\citep{metropolis1953equation, kirkpatrick1983optimization, del2006sequential}. Our approach is of this second type, though it is important to note that a solution from our method could be used as a warm-start for one of the MIP solvers; a better warm-start is the key to finding optimal solutions faster with MIP.

There are two main steps per iteration in these types of algorithms: (i) coordinate descent steps involving a line search along the objective function, often using a local surrogate function, and (ii) local swaps, where the support set (the set of features permitted to have nonzero coefficients) changes over iterations. Our work advances both of these steps over previous work. For (i), we show that a natural surrogate for the logistic loss used in previous work leads to inefficiency, in that its step sizes are provably too conservative. We propose a more aggressive step. This opens up the possibility of using \textit{cutting planes} or \textit{quadratic cuts}. Cuts often help us rapidly prune the search space: by comparing the lower bound from the cuts with the current best loss, we are often able to prove that there is no possible step size we could take that would reduce our objective, in which case we will try a more promising direction in the search space. 
The $\ell_2$ penalty term permits us to use quadratic cuts. When we do not want the $\ell_2$ term (i.e., $\lambda_2=0$), we can use cutting planes.
For (ii), we find that the order in which we evaluate features plays an important role, which has been previously overlooked. We use a priority queue to dynamically manage the order of evaluating features.
The priority queue discourages us from checking features that are unlikely to change the model's support set, making the process of finding high-quality solutions more efficient.

In addition, for (i), improving the speed of the coordinate descent steps, we propose to use the exponential loss, which has a major advantage over the logistic loss in that the line search taken at each coordinate descent iteration has an analytical solution. Another appealing property of the exponential loss is that its probabilistic interpretation is extremely similar to that of logistic regression.
Also, minimizing the exponential loss is known to provably maximize a proxy for the Area Under the ROC Curve \citep{ErtekinRu11}, making it an ideal choice for this problem. 

Our contributions are:
\begin{enumerate}
\item We prove that previous work on surrogate loss optimization yields step sizes that are too conservative (Theorem \ref{theorem:limitedCDMove}). 
\item When $\lambda_2 = 0$, we propose a linear cutting plane algorithm that prunes the search space by efficiently determining whether adding a feature could potentially reduce the objective.
\item With a small amount of $\ell_2$ regularization, we propose a quadratic cut algorithm giving a tighter lower bound than the linear cutting plane algorithm.
\item We propose a method using the exponential loss, which is cleaner and simpler.
\item For more efficient best subset search, we use a priority queue to dynamically manage the order of checking features.
\end{enumerate}

Our algorithms provide a dramatic improvement over previous approaches, often achieving the same results in less than half the time, and are able to produce models for thousands of features and observations in seconds. For instance, on the challenging FICO dataset from the 2018 Explainable Machine Learning Challenge, which, after the transformation to dummy variables, has 1,917 dummy features and 10K observations, we produce a generalized additive model of 19 total dummy variables, with performance comparable to black-box performance, in under 5 seconds.

%% file: sections/02-background.tex
\section{BACKGROUND} 
Coordinate descent is popular in machine learning. Other techniques that use variations of it include AdaBoost \citep{freund1997decision} and Sequential Minimal Optimization used for support vector machines \citep{platt1998sequential}. 
Surrogate functions are also common, e.g., they are used by Expectation Maximization \citep{dempster1977maximum}. 
We begin with background, following \citet{patrascu2015random} and \citet{dedieu2020learning}.

The loss function in Problem \eqref{l0_l2_loss} can be rewritten as:
\begin{align*}
    \mathcal{L}(\bw) & = G(\bw) + \lambda_{0} \lVert \bw\rVert_{0},
\end{align*}
with $G(\bw) = \sum_{i=1}^n \log(1+\exp(-y_i (\bx_i^{T} \bw))) + \lambda_2 \lVert \bw\rVert_2^2$.

Let us optimize $\mathcal{L}(\bw)$ along coordinate $j$ starting at point $\bw^t$ at iteration $t$.
Let $\nabla_j G(\bw^t)$ denote the $j$-th component of the gradient of $G(\bw^t)$,  and let $L_j$ be the Lipschitz constant for $\nabla_j G(\bw^t)$. For any $d \in \mathbb{R}$:
\begin{align*}
    |\nabla_j G(\bw^t + \be_j d) - \nabla_j G(\bw^t)| \leq L_j |d|
\end{align*}
where $\be_j$ is a vector with all components equal to $0$ except for the $j$-th component, which is equal to $1$.
A surrogate upper bound on $G(\bw^t + \be_j d)$ is thus:
\begin{align}
    G(\bw^t+\be_j d) \leq G(\bw^t) + d \nabla_j G(\bw^t) + \frac{1}{2} L_j d^2.
\end{align}
Instead of minimizing the original loss function with respect to coordinate $j$ (as would be typical in coordinate descent), we can minimize this quadratic upper bound with the new coefficient $w^{t+1}_j = w^t_j + d$:
\begin{align*}
    \hat{w}_j^{t+1} \in& \argmin_{u} \;G(\bw^t) + (u - w_j^{t}) \nabla_j G(\bw^t) \\ 
    & + \frac{1}{2} L_j (u - w_j^{t})^2 + \lambda_0 \mathbb{1}_{u \neq 0}.
\end{align*}
Following previous work~\citep{dedieu2020learning}, we have an analytical solution for the above problem:
\begin{align}
\label{formula:thresholding}
    \hat{w}_j^{t+1} = T(j, \bw) = \begin{cases}
			c, & \text{if } |c| \geq \sqrt{\frac{2\lambda_0}{L_j}}\\
            0, & \text{otherwise}
		 \end{cases}
\end{align}
where $c = w_j^t - \nabla_j G(\bw^t)/L_j$.

If a solution cannot be improved by coordinate descent using this surrogate and thresholding function, we say this solution is \textit{surrogate 1-OPT}, meaning that no single coordinate can be changed to improve the objective when using this surrogate for the line search.



As discussed earlier, local swap, add, and remove operations are useful for best subset search and other local search problems.
These govern the support of the coefficient vector, determining which coefficients are permitted to be nonzero.
We use $S$ to denote the support of the feature vector; that is, the set of features that are permitted to have nonzero coefficients.
We can swap some features in the current support, denoted by $S_1 \subseteq S$, for other features not in the support, denoted by $S_2 \subseteq S^c$.
After each swap, we optimize the coefficients that are permitted to be nonzero. 

To reduce computational cost, while evaluating a possible swap, we use an approximate evaluation procedure where we update only the coefficients of the swapped features and keep coefficients of other unswapped features fixed. If such a swap leads to a better loss, we add $S_2$ to the support, remove $S_1$ from the support, and update all coefficients for the features in the new support. 
We will focus on single feature swaps (i.e. $|S_1| = |S_2| = 1$) in this work.
If no allowed swap appears to improve the loss, then we call the solution a \textit{swap 1-OPT} solution.

%% file: sections/03-overview.tex
\section{OVERVIEW OF FAST SPARSE LOGISTIC REGRESSION}\label{sec:overview}

\begin{figure*}[t]
    \vspace{-2mm}
    \centering
    \includegraphics[width=.9\textwidth]{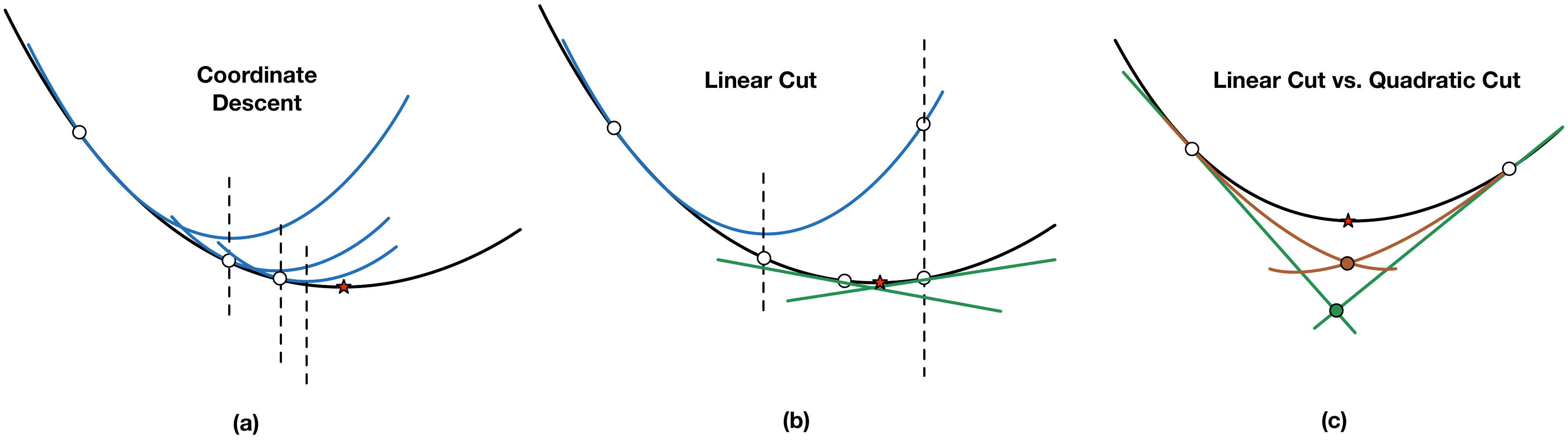}
    \caption{
    (a) We repeatedly apply coordinate descent until convergence to get the optimal coefficient (shown by the red star) and then calculate the loss. (b)  We calculate a lower bound of the optimal loss by constructing two cutting planes. We can rule out the new feature if the lower bound of the loss from the cutting planes is larger than the best current loss. (c) Quadratic cuts (in red) form the lower bound instead and are tighter.}
    \vspace{-2mm}
    \label{fig:linear_cut}
\end{figure*}

Let us focus on the logistic loss.
Given an initial solution, we optimize one feature's coefficient at a time, and swap features within the support set to improve the solution.
Our technique evaluates whether it could be worthwhile to swap two features. It is based on a theorem showing that thresholding from \eqref{formula:thresholding} yields step sizes that are too conservative. Using this information, we develop an algorithm that uses \textit{quadratic cuts}. Typically, cutting planes \citep{kelley1960cutting} are used in mathematical programming solvers, whereas here, we use cuts as part of efficient feature elimination within coordinate descent. 
Our second technique uses a priority queue to manage the search order for pairs of features to swap.
At each outer iteration, we drop a feature $j$ in the support and at each inner iteration, we evaluate adding a feature $j'$. The full pseudocode is in Appendix~\ref{app:pseudocode}. The main steps are:

1. \textbf{Remove and find alternatives.} According to the priority queue, try removing feature $j$ from the current support. Find $J'$ features outside the support as alternatives for feature $j$. These alternative features are picked according to orthogonal matching pursuit~\citep{lozano2011group}. For each feature $j'\in J'$, we evaluate whether it is worthwhile to include it in our support as a replacement of feature $j$. This is done using the following procedures.

2. \textbf{Aggressive step.} Given a new feature $j'$ that we may want to include in our support, we wish to find two values on  opposite sides of the optimal coefficient $w_{j'}^*$. However, at current value $w_{j'}$, the thresholding results stay on a single side of the optimal value (as we will prove in Theorem \ref{theorem:limitedCDMove}). Thus, we take an aggressive step by going double the distance suggested by thresholding, or triple the distance if necessary. If this triple-sized step does not get to the opposite side of $w_{j'}^*$, we iteratively apply thresholding  \eqref{formula:thresholding} to get a near-optimal coefficient and move to Step 6.


3. \textbf{Binary search.} Suppose we have found two values $a$ and $b$ on opposite sides of $w_{j'}^*$. We then perform one binary search step to get a point closer to $w_{j'}^*$ by setting $c$ to be the midpoint, $c = \frac{1}{2}(a+b)$. If $c$ is on the same side of $a$, we replace $a$ with $c$; if not, we replace $b$ with $c$.
We use quadratic cuts (via the \textit{Quadratic Cut Bound}, Theorem \ref{eq:quadratic_cut_bound}) at points $a$ and $b$ to obtain a lower bound on the objective for the optimal coefficient of the feature. In the case of no $\ell_2$ regularization, we use cutting planes instead. More detail on this is in the next section.

4. \textbf{Eliminate.} If the lower bound is larger than the current best loss we have encountered so far, the new feature can be eliminated from consideration; we do not add this feature into our support. We move onto the next possible feature and start again from Step 2.

5. \textbf{Line search.} If the lower bound is smaller than the current best loss we have encountered, then feature $j'$ could lead to a better solution. Therefore, we iteratively use thresholding  \eqref{formula:thresholding} to obtain a near-optimal coefficient for the line search. (Alternatively, we could continue binary search for the minimum.)  

6. \textbf{Complete the step.} We then calculate the loss with respect to this near-optimal coefficient for the line search. If the loss is higher than the current best loss, we eliminate this feature and move to the next best alternative feature; if the loss is lower, we add this new feature $j'$ into the support to make up for the removed feature $j$ and optimize all of the coefficients, completing a successful swap step.

7. \textbf{Update priority queue.} If no alternative feature can replace feature $j$, we add feature $j$ back into the support and rate feature $j$ less promising in our priority queue. This allows us to explore features that have a better chance of being swapped with an alternative feature next time.

%% file: sections/04-cuts.tex
\section{SURROGATE QUADRATIC CUTS}\label{sec:quadcuts}
Let us provide the theorem motivating our coordinate descent method for the logistic loss, which shows that the step sizes from thresholding in \eqref{formula:thresholding} are too conservative.
Recall that thresholding is derived by minimizing a quadratic upper bound of the loss function.
The coefficient of the quadratic function is the Lipschitz constant, which defines the maximum curvature the loss function can achieve.
These connections imply:
\begin{theorem}(Thresholding is too conservative.)  \label{theorem:limitedCDMove}
Let $\bw^{t}$ be the current solution at iteration $t$, $w^t_j$ be the coefficient for the $j$-th feature, and let $w_j^*$ be the optimal value on the $j$-th coefficient while keeping all other coefficients fixed to their values at time $t$.
Furthermore, let $\bw^{t+1}=\bw^t + \be_j (T(j, \bw^t) - w_j^t)$, where $\be_j$ is a vector with $1$ on the $j$-th component and 0 otherwise and $T(j, \bw^t)$ is the thresholding operation with the support set fixed (i.e., $\lambda_0=0$).
Then we have the following inequalities:
\begin{align}\label{Th1:1}
    \nabla_{j}G(\bw^t) \nabla_{j} G(\bw^{t+1}) \geq 0, \\\label{Th1:2}
    (w_j^{t} - w_j^*)(w_j^{t+1} - w_{j}^{*}) \geq 0, \\ \label{Th1:3}
    \text{ and } G(\bw^t) \geq G(\bw^{t+1}).
\end{align}
\end{theorem}
This result shows that the thresholding operation will move the coefficient of the $j$-th feature closer to the optimal value $w_{j}^{*}$ with a smaller loss value, as shown by \eqref{Th1:3}. However, the coefficients before and after the thresholding operation will \textit{always remain on the same side of} $w_{j}^{*}$, as shown by either \eqref{Th1:2} or \eqref{Th1:1}. To see this, consider \eqref{Th1:2}. We have two scalars of the same sign: $w_j^t - w_j^*$ and $w_j^{t+1} - w_{j}^{*}$. If $w_j^{t+1}$ were on the opposite side of $w^*$ than $w_j^t$, the product of these two scalars would instead be negative. Alternatively, by \eqref{Th1:1}, if the slope of $G$ at $\bw^t$ is negative, the slope at $\bw^{t+1}$ is also negative, indicating that we have not yet passed the minimum (of our convex logistic loss).
Thus, this theorem indicates that the step size provided by the surrogate is too conservative; the distance is always too small to reach $w_j^*$. Figure \ref{fig:linear_cut} (left) illustrates this issue. The algorithm may make several steps before becoming sufficiently close to $w_j^*$.

\textit{Our technique chooses an aggressive step size that takes us beyond $w_j^*$, in order to use \textit{cuts} to produce a lower bound on the loss at $w^*$.} If the lower bound is too high, we can exclude the feature all together.

The first type of cut we introduce is classical \textit{cutting planes}, which provide a linear lower bound on the loss. This can be used even if we have only $\ell_{0}$ regularization on the logistic loss (i.e., if $\lambda_2$ in \eqref{l0_l2_loss} is 0).
With an additional $\ell_{2}$ penalty term, we can obtain a strictly tighter lower bound on the loss, yielding quadratic cuts. We introduce both types of cuts next, starting with cutting planes.


\begin{theorem}(\textit{Classical cutting planes, not novel to this paper}) \label{eq:linear_cut_bound}
Suppose $f(x)$ is convex and differentiable on domain $\mathbb{R}$.
Let $\alpha_1$ and $\alpha_2$ be slopes of tangent lines of $f(x)$ at locations $x_1$ and $x_2$.
If $\alpha_1 \alpha_2 \leq 0$, there is a lower bound on the optimal value $f(x^*)$:
\begin{align}
\label{eq:linear_cut_twoPoints}
    f(x^*) \geq \frac{\alpha_1 f(x_2) - \alpha_2 f(x_1) + \alpha_1 \alpha_2 (x_1 - x_2)}{\alpha_1 - \alpha_2}.
\end{align}
\end{theorem}
This method originates from a first-order approximation of function $f(x)$. Figure~\ref{fig:linear_cut}(b) shows linear cuts.

With an additional $\ell_{2}$ penalty term, we can obtain a strictly tighter lower bound on the loss via \textit{quadratic cuts}. The  $\ell_{2}$ term makes $G(\bw)$ strongly convex, which means for any two points $\bw$ and $\bw'$ in the domain: 
\begin{align*}
    G(\bw') \geq G(\bw) + \nabla G(\bw)^{T} (\bw'-\bw) + \lambda_{2} \lVert \bw' - \bw\rVert_2^2.
\end{align*}
Using this strongly convex property, we can tighten the lower bound given in Theorem~\ref{eq:linear_cut_bound} as follows:
\begin{theorem}(Quadratic Cut Bound) \label{eq:quadratic_cut_bound}
Suppose $f(x)$ is strongly convex and differentiable over $\mathbb{R}$ with $\lambda_2$ for the coefficient of the quadratic term.
Let $\alpha_1$ be the slope of the tangent line to $f(x)$ at location $x_1$. Then, there is a lower bound on the optimal value $f(x^*)$:
\begin{align}
    \label{eq:quadratic_cut_onePoint}
    f(x^*) \geq \mathcal{L}_{\textrm{low}}:=f(x_1) - \frac{\alpha_1^2}{4 \lambda_2}.
\end{align}
Let $\alpha_2$ be the slope of the tangent line to $f(x)$ at another location $x_2$.
If $\alpha_1 \alpha_2 \leq 0$, a lower bound on the optimal value $f(x^*)$ is as follows:
\begin{equation}
    \label{eq:quadratic_cut_twoPoints}
    f(x^*) \geq \mathcal{L}_{\textrm{low}}:=f(\hat{x}) + \alpha_1 (\hat{x} - x_1) + \lambda_2 (\hat{x} - x_1)^2,
\end{equation}
\[
     \hat{x} = \frac{-f(x_1) + f(x_2) + \alpha_1 x_1 - \alpha_2 x_2 - \lambda_2 (x_1^2 - x_2^2) }{\alpha_1 - \alpha_2 - 2\lambda_2 (x_1 - x_2)}. \nonumber
\]
\end{theorem}
Since this method originates from a second-order approximation of the function $f(x)$, we name this bound the Quadratic Cut Bound. Either this bound or cutting planes helps us decide when not to include a potential feature in our support, even without knowing its optimal coefficient from the line search.

%% file: sections/05-exponential.tex
\section{FAST SPARSE CLASSIFICATION WITH EXPONENTIAL LOSS}\label{sec:exp_loss}
Let us now switch from logistic loss to the exponential loss,
optimizing:
\[
	\min_{\bw}\left[ \sum_{i=1}^{n} \exp(-y_i\bw^T\bx_i) + \lambda_{0} \lVert \bw \rVert_{0}\right].
\]
Though exponential loss typically is not used for sparse classification, it has no clear disadvantages over the logistic loss and even has several advantages. 
First we point out that exponential loss and logistic loss have remarkably \textit{similar probabilistic interpretations} under the assumption that we have captured the correct set of features.  
While logistic regression estimates conditional probabilities as 
$\hat{P}_{\textrm{logistic}}(y=1|\bx) = \frac{e^{f(\bx)}}{1+e^{f(\bx)}}$
where $f(\bx) = \bw^T \bx$,
the exponential loss has almost the same probabilistic model:
$\hat{P}_{\textrm{exp loss}}(y=1|\bx) = \frac{e^{2f(\bx)}}{1+e^{2f(\bx)}}.
$
Thus, both loss functions are equally relevant for modeling conditional probabilities.

The main benefit of exponential loss is that it has an \textit{analytical solution for the line search} at each iteration when features are binary ($-1$ and $1$). This avoids the necessity for cutting planes, quadratic cuts, or even surrogate upper bounds. Following the derivation of AdaBoost as a coordinate descent method \citep{schapire2013boosting}, its line search solution follows the formula $\frac{1}{2}\ln\left(\frac{1-d_-}{d_-}\right)$, where $d_-$ indicates the weighted misclassification error of the feature chosen at iteration $t$ (here we are interpreting each weak classifier as an individual feature, and the weak learning algorithm picks one of these features per iteration). AdaBoost's weight update step avoids calculation of the exponential loss at each iteration, and the full procedure is extremely efficient. 
(The main difference between our method and this reduced version of AdaBoost is that AdaBoost is not designed to yield sparse models.) In the following theorem, we provide a condition under which our method would decline to add a new feature at iteration $t$, because it does not provide an overall benefit to our objective. 
We use $\bz_i \in \mathbb{R}^{p}$ with $\bz_{i} = y_i \bx_{i}$ to succinctly represent the product between $y_i$ and $\bx_i$. The objective can be then rewritten as:
\begin{align*}
    \min_{\bw} [H(\bw) + \lambda_0 \lVert \bw \rVert_0]
\end{align*}
where $H(\bw) = \sum_{i=1}^n \exp(-\bw^T \bz_i)$.

\begin{figure*}[ht]
    \centering
    \includegraphics[width=\textwidth]{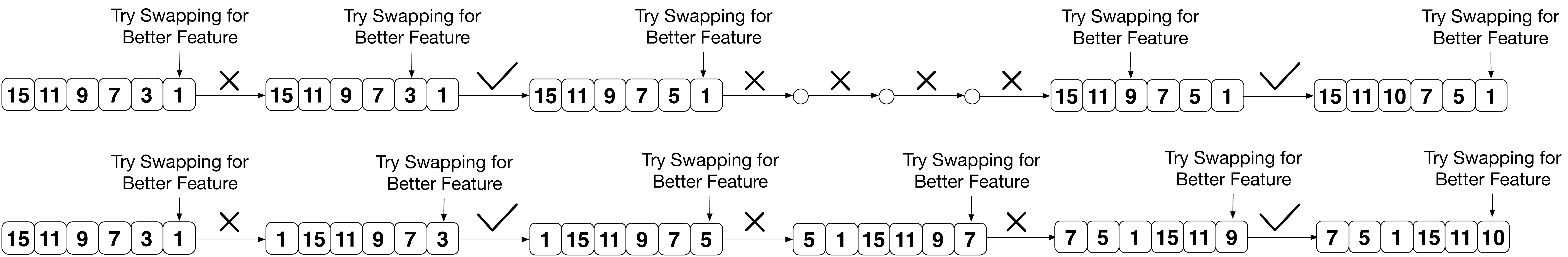}
    \caption{Sequential Ordering vs$.$ Dynamic Ordering. Upper: We check each feature sequentially. Whenever we find a better feature, we always start from the beginning to find the next possible swap. Lower: We order the list, checking the feature that has failed the least amount of times first.
    We hold off checking less promising features until the end, saving substantial computational time. 
    }
    \label{fig:dynamic_ordering}
\end{figure*}

\begin{theorem}\label{thm:exp_loss}
Let $\bw^t$ be the coefficient vector at iteration $t$, $H^t:=H(\bw^t)$ and $\lambda_0$ be the regularization constant for the $\ell_0$ penalty. For the $j$-th coordinate, we update the coefficient according to:

(1) Suppose $w_j^t \!= \!0$. Let $d_- \!=\! \sum_{i: z_{ij}=-1}\!c_i / \!\sum_{i = 1}^n \! c_i$, with $c_i = \exp(-(\bw^t)^T \bz_i)$. If $d_-$ is within the interval:
\[
\left[\frac{1}{2}\!-\!\frac{1}{2H^t}\sqrt{\lambda_0(2H^t\!-\!\lambda_0)},\frac{1}{2}\!+\!\frac{1}{2H^t}\sqrt{\lambda_0(2H^t\!-\!\lambda_0)}\right],
\]
then set $w_{j}^{t+1}$ to 0. Otherwise set $w_{j}^{t+1} = \frac{1}{2}\ln \frac{1-d_-}{d_-}$.

(2) Suppose $w_j^t \!\neq\! 0$.  Let $D_- \!=\! \sum_{i: z_{ij}=-1}\!c_i / \!\sum_{i = 1}^n \! c_i$, with $c_i = \exp(-(\bw^t - w^t_j \be_j)^T z_i)$. Let $H^t_{\neg j} = H(\bw^t-w^t_j \be_j)$. 
If $D_-$ is within the interval:
\[
\left[\frac{1}{2}\!-\!\frac{1}{2H^t_{\neg j}}\sqrt{\lambda_0(2H^t_{\neg j}\!-\!\lambda_0)},\frac{1}{2}\!+\!\frac{1}{2H^t_{\neg j}}\sqrt{\lambda_0(2H^t_{\neg_j}\!-\!\lambda_0)}\right],
\]
then set $w_j^{t+1}$ to 0. Otherwise, set $w_j^{t+1}=\frac{1}{2}\ln \frac{1-D_-}{D_-}$.
\end{theorem}

Another potential benefit of the exponential loss is that it is a surrogate for the AUC, i.e., Area Under the ROC Curve \citep[]{ErtekinRu11}. Thus, we have reason to expect good AUC performance when optimizing the exponential loss.

%% file: sections/06-dynamic.tex
\section{DYNAMIC FEATURE ORDERING}

 Now that we can optimize along the coordinates using either logistic loss (Sections \ref{sec:overview} and \ref{sec:quadcuts}) or exponential loss (Section \ref{sec:exp_loss}), we discuss the important swap steps that help the algorithm drop features that have promising swap candidates. As stated in Section \ref{sec:overview}, after coordinate descent is run until a local minimum is reached, we alternate between coordinate descent steps and swap steps. The technique proposed here is broadly applicable and can improve the speed not only for the logistic loss and the exponential loss but also for the squared loss in linear regression (see Appendix~\ref{app:linear_regression}).

We focus on the swap 1-OPT solutions (i.e., $|S_1| = |S_2| = 1$). The order of checking features in $S_1$ for possible swaps is key to improving the efficiency. Instead of checking features in $S_1$ sequentially based on feature indices \citep{dedieu2020learning}, we dynamically order these features via a priority queue. We provide an example in Figure~\ref{fig:dynamic_ordering} to illustrate the key difference between the two approaches.

Suppose we have an initial solution with support on features $1, 3, 7, 9, 11, \text{and } 15$, and features 3 and 9 are suboptimal.
We can swap feature 3 with feature 5 and feature 9 with feature 10 to get a lower total loss.
The first method checks features sequentially and always starts from the first index in the support after a successful swap.
The algorithm terminates if we have checked all features without making any swaps.
This method implicitly assumes that each feature in the support  has an equal probability of having a successful swap. However, a feature that has not been swapped for many iterations is likely to be important and therefore unlikely to be swapped in the near future.
It is better to check more promising features first.

To achieve this, we record how many times a feature has failed to swap. The features are ranked in ascending order of the number of failure times. Features that have never been checked are kept at the top of our priority queue.
This local search process terminates when all features have been evaluated (i.e., the full priority queue) without making a successful swap.
This accelerates the process to reach a swap 1-OPT solution.


%% file: sections/07-experiment.tex

\section{EXPERIMENTS}
Our evaluation answers the following questions: 
(1) How well do our early pruning technique, priority queue ordering, and proposed exponential loss perform in terms of run time relative to the state-of-the-art? (\S \ref{sec:comp_eff}) (2) How well do our methods perform in terms of AUC, accuracy, and sparsity relative to state-of-the-art algorithms on simulated and real datasets? (\S \ref{sec:exp_qual})

We compare our methods to $\ell_1$ regularized logistic regression (LASSO) via the \textit{glmnet} package \citep{glmnet}, MCP via the \textit{ncvreg} package \citep{ncvreg}, and L0Learn \citep{dedieu2020learning}.
We use the fast C\texttt{++} linear algebra libraries of L0Learn in our implementation. For all datasets, we run 5-fold cross validation and report the mean and standard deviation. 
Appendix \ref{app:exp_details} presents the experimental setup, datasets, and evaluation metrics, and Appendix \ref{app:exp_results} presents additional experimental results. 
Our methods are denoted as LogRegQuad-L0 (logistic loss and quadratic cuts) and Exp-L0 (exponential loss).


\begin{figure*}[t]
  \vspace{-3mm}
  \centering
  \includegraphics[width=1.0\linewidth]{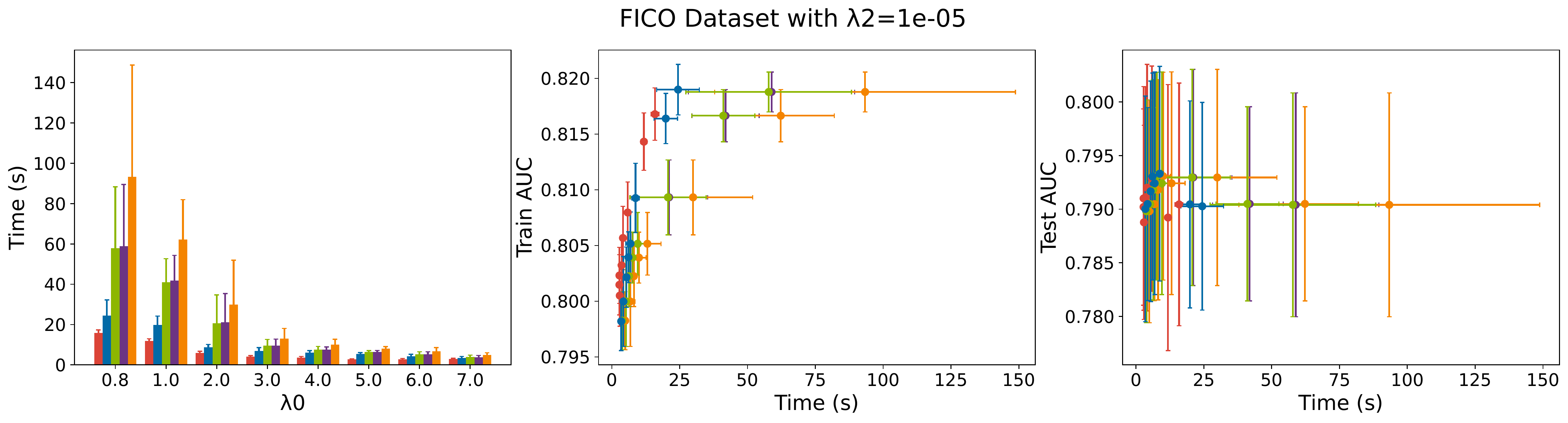}
  \includegraphics[width=1.0\linewidth]{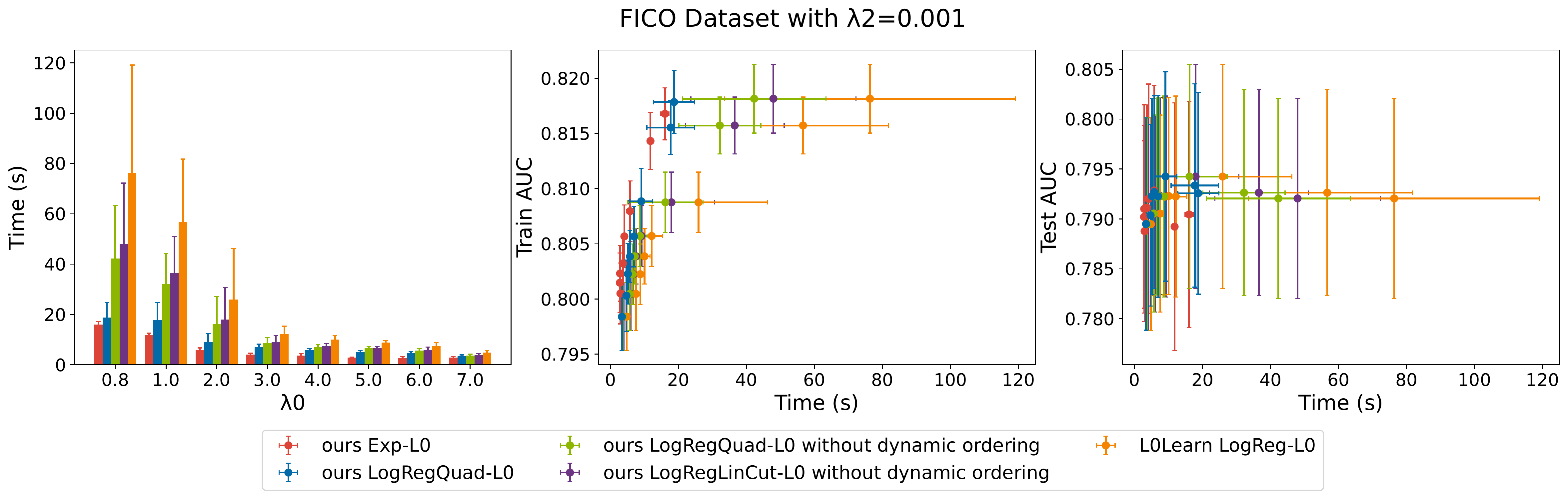}
  \caption{Computational times of different methods. ``Exp'' stands for exponential loss, ``LogReg'' stands for logistic loss, ``LinCut'' stands for linear cuts, and ``Quad'' stands for quadratic cuts. Note that there is no $\ell_2$ penalty for the exponential loss. \textit{Our Exp-L0 method is generally about 4 times faster than L0Learn.} Note that the AUC axes indicate practically similar performance for these particular methods; the training time is what differentiates the methods. Additionally, when the $\ell_2$ penalty increases from $\lambda_2=1\text{e}-05$ to $\lambda_2=0.001$, there is a computational speedup from using the linear cut to the quadratic cut due to the tighter lower bound.}
  \label{fig:time_speed_up}
\end{figure*}

\begin{figure*}[t]
  \centering
  \includegraphics[width=1.0\linewidth]{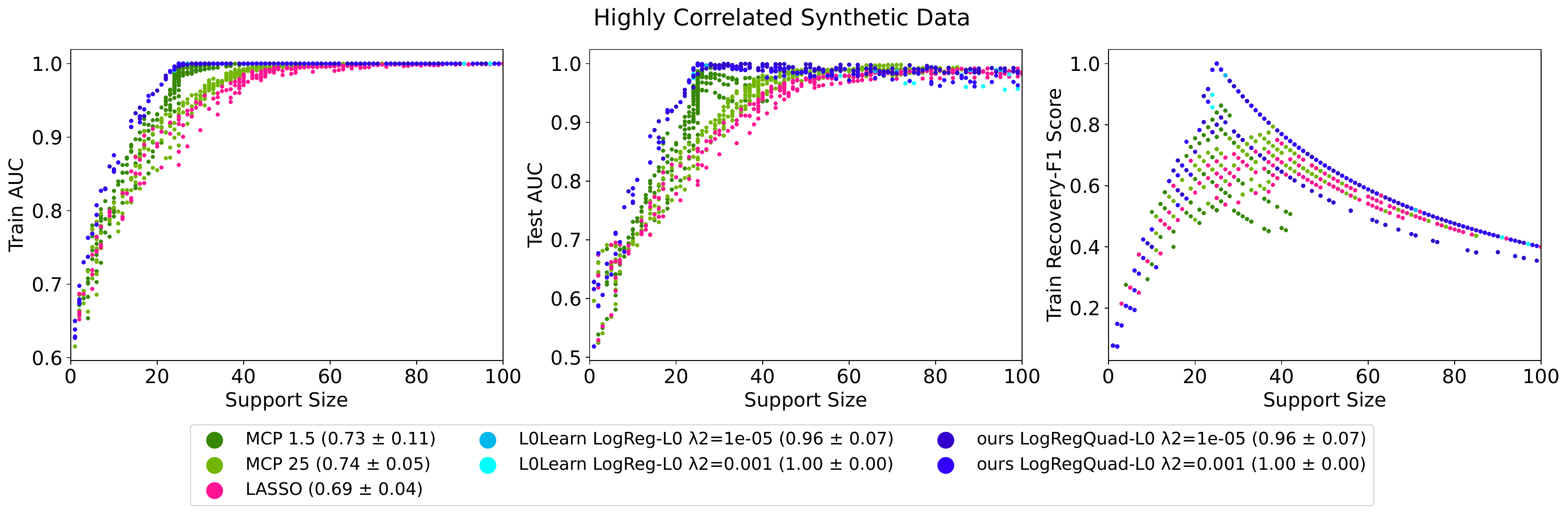}
  \caption{Results from all 5 datasets (each dataset generated by a different random seed) and parameter choices on highly correlated synthetic datasets.
  The parentheses contain the best Recovery-F1 scores averaged over all 5 datasets. MCP is shown with $\gamma$ fixed at 1.5 and 25, and all other choices for $\gamma$ lie between the shown regions.
  Our methods and L0Learn outperfom  MCP and LASSO 
  in terms of the AUC  (left and middle), and better recover the true support (right). 
  L0Learn's performance heavily overlaps with our methods. Our methods have a computational advantage over L0Learn as shown in the last section.}
  \label{fig:synthetic_data}
\end{figure*}

\begin{figure*}[!ht]
  \centering
  \includegraphics[width=1.0\textwidth]{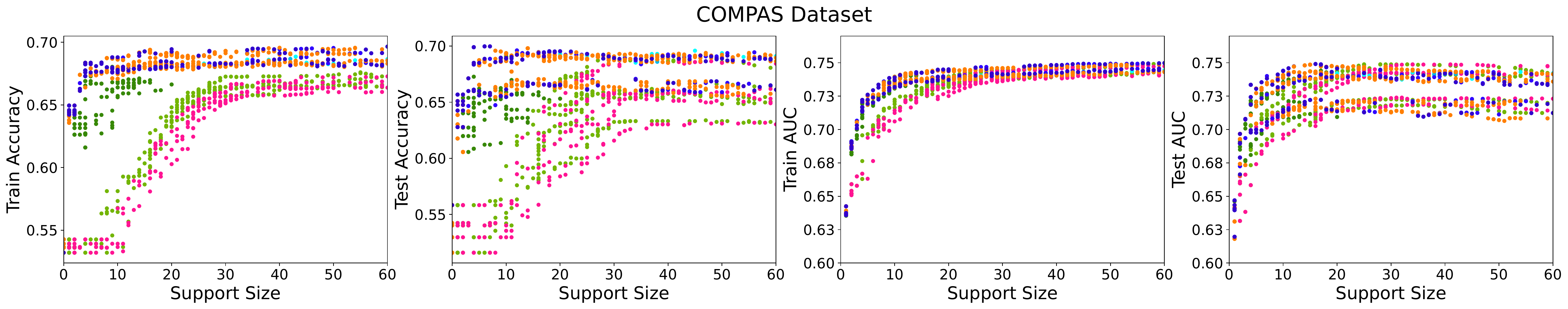}
  \includegraphics[width=1.0\textwidth]{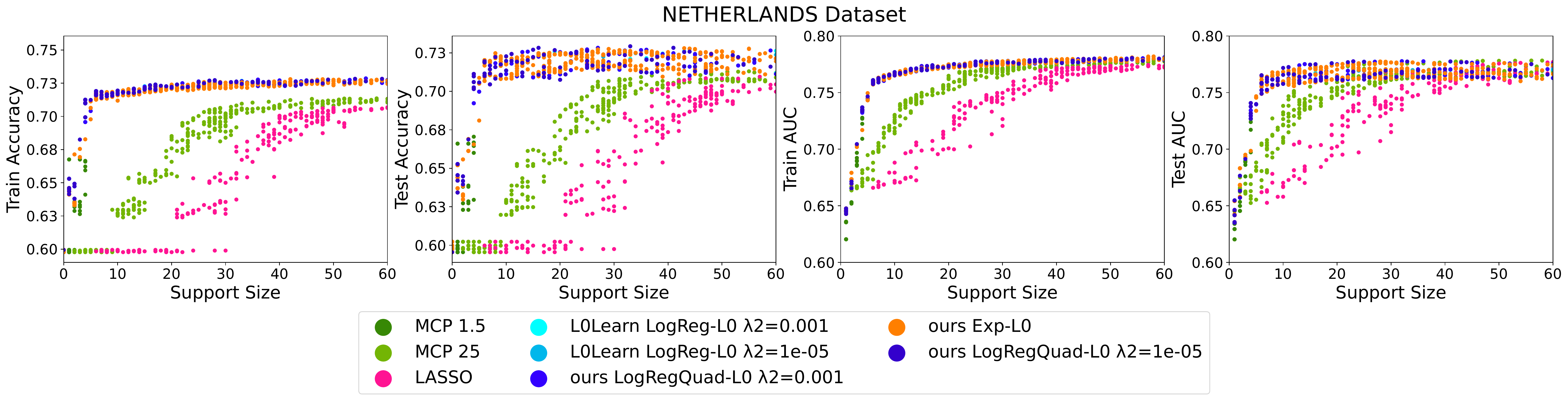}
\caption{Results from all folds and parameter choices on real datasets: COMPAS and NETHERLANDS. We can see from the first and second columns (training and test accuracies) that MCP and LASSO do not perform well. Our methods and L0Learn (overlapping) outperform all other methods. Our methods are more computationally efficient than L0Learn.}
\label{fig:real_data}
\end{figure*}

\subsection{Computational Efficiency}\label{sec:comp_eff}
To examine the impact of the quadratic cuts and dynamic ordering, we first run our algorithm with only quadratic cuts and then enable dynamic ordering on the FICO dataset from the Explainable Machine Learning Challenge \citep{fico}. We also run this experiment using Exp-L0. L0Learn is used as a baseline. (MCP and LASSO use continuous regularization terms, which provides them with a run-time advantage, though these methods do not perform as well, as shown in the next subsections.)
The $\ell_{0}$ parameters we used are $\{0.8, 1, 2, 3, 4, 5, 6, 7\}$ and the $\ell_{2}$ parameters used are $\{0.00001, 0.001\}$. 

Figure \ref{fig:time_speed_up} shows the training time and AUC values on the FICO dataset. The methods achieve performance comparable with \citet{ChenEtAl21}, who reported best black-box AUC $\sim$0.8. 
Our method using only linear cuts (\textcolor{purple}{purple} bars) runs faster than the baseline (\textcolor{orange}{orange} bars, L0Learn) for all regularization options.
With $\ell_2$ regularization coefficient $\lambda_2 = 0.001$, the time is reduced when we switch from using linear cuts to quadratic cuts (\textcolor{ForestGreen}{green} bars) due to the tighter lower bound, as in Figure~\ref{fig:linear_cut}.
The training time is further reduced by using both quadratic cuts and dynamic ordering (\textcolor{blue}{blue} bars, which is LogRegQuad-L0). 
\textit{Exp-L0 (\textcolor{red}{red} bars) is the fastest approach}.
Again, this speed-up owes to the analytical line search and fast update.

From the four rightmost subfigures, we find that our improvement in training time does not negatively impact training/test AUC scores, as our methods (\textcolor{red}{red} and \textcolor{blue}{blue} dots) form a ``left frontier'' with respect to the baseline L0Learn (\textcolor{orange}{orange} dots). Results for additional datasets are in Appendix \ref{app:run_time_savings}.

\subsection{Solution Quality}\label{sec:exp_qual}
We next evaluate sparsity vs$.$ performance. In addition to AUC on the datasets, we calculate Recovery-F1 score to measure how well we captured the ground truth support (ground truth coefficients ${\bw}^*$ are known for simulated datasets). Recovery-F1 score is $\frac{2PR}{P+R}$, where $P = |\text{supp}(\hat{\bw}) \cap \text{supp}(\bw^*)| / |\text{supp}(\hat{\bw})|$ is the precision and $R = |\text{supp}(\hat{\bw}) \cap \text{supp}(\bw^*)| / |\text{supp}({\bw}^*)|$ is the recall. $\text{supp}(\cdot)$ stands for the support (indices with nonzero coefficients) of a solution. We can use Recovery-F1 score for synthetic data only, since we need to know ${\bw}^*$ to calculate it. 

\textbf{Synthetic Data:} Figure~\ref{fig:synthetic_data} shows sparsity/AUC tradeoffs and sparsity/Recovery-F1 tradeoffs on a synthetic dataset consisting of highly correlated features. 
Our methods are generally tied for the best results. LASSO (\textcolor{Magenta}{pink} curves) and MCP (\textcolor{ForestGreen}{green} curves) do not fully optimize the AUC, nor recover the correct support. 
For the full regularization path, the AUC's of L0Learn and our method largely overlap. However, as demonstrated in the previous subsection, our method runs much more quickly than L0Learn.

Since the features for this synthetic dataset are continuous (and we chose not to binarize them), Exp-L0 cannot be applied; its advantage comes from exploiting its analytical line search for binary features. 

\textbf{Real Datasets:} Figure~\ref{fig:real_data} shows sparsity-AUC tradeoffs and sparsity-accuracy tradeoffs on the COMPAS and NETHERLANDS datasets. 
LASSO and MCP do not achieve high prediction accuracy on training and test sets.
L0Learn and our proposed methods have higher AUC and accuracy. Again, while L0Learn and our methods are tied for the best performance (which could be the optimal possible performance for this problem), our methods have major advantages in speed.
More results are in Appendix \ref{app:support_tradeoffs}.

%% file: sections/08-relatedWorks.tex
\section{RELATED WORK}
\textbf{Mixed Integer Optimization.}
There have been many approaches to finding the optimal solution to logistic regression either with an $\ell_{0}$ regularization or cardinality constraint~\citep{sato2016feature, sato2017piecewise, ustun2017optimized, bertsimas2017logistic, bertsimas2017sparse, sakaue2019best, ustun2019learning}.
In general, these approaches formulate the problem as a mixed-integer optimization problem~\citep[see][]{bertsekas1997nonlinear, wolsey1999integer}.
The problem can then be solved using branch-and-bound search~\citep[see][]{land2010automatic} or cutting-plane methods~\citep{kelley1960cutting, gilmore1961linear, gilmore1963linear}.
However, even with the recent advances in hardware and software, MIP solvers are orders of magnitude slower than the methods we consider here and requires relatively large $\ell_2$ regularization to work well~\citep{bertsimas2017sparse, dedieu2020learning}.

\textbf{Gradient-based Heuristic Methods.}
One of the most widely used methods to promote sparsity is LASSO~\citep{tibshirani1996regression}, which relaxes the $\ell_{0}$ penalty to $\ell_{1}$.
However, $\ell_{1}$ simultaneously promotes sparsity and shrinks the coefficients, leading to bias.
Several new methods obtain solutions  under cardinality constraints or $\ell_{0}$ penalty terms.
One method is Orthogonal Matching Pursuit (OMP)~\citep{lozano2011group, elenberg2018restricted}, which greedily selects the next-best feature based on the current support and gradients on coefficients.
Other methods include Iterative Hard Thresholding (IHT)~\citep{blumensath2009iterative}, coordinate descent~\citep{beck2013sparsity, patrascu2015random, dedieu2020learning}, GraSP~\citep{bahmani2013greedy}, and NHTP~\citep{zhou2021global}.
These methods enjoy fast computation, but their solutions suffer when the feature dimension is high or features are highly correlated because they can get stuck at local minima~\citep{dedieu2020learning}.

\textbf{Local Feature Swaps.} 
Some recent work considers swapping features on a given support.
One such example is ABESS~\citep{zhu2020polynomial, zhang2021certifiably}, which 
ranks features based on their contribution to the loss objective.
Then, they swap only unimportant features in the support with features outside the support.
Our experiments show that ABESS often returns ``nan'' values for its coefficients, thus in its current form was not able to be included in our experiments.
Another work is L0Learn~\citep{hazimeh2020fast, dedieu2020learning}, which exhaustively tries replacing every feature in the support with better features.

To the best of our knowledge, our work is the first where quadratic cuts (or exponential loss) and dynamic ordering have been used for sparse classification.

%% file: sections/09-conclusion.tex
\section{CONCLUSION}
We have shown substantial speedups over other techniques for best subset search for probabilistic models with high-quality solutions. Our advances are due to several key ideas: (1) the use of cutting planes and quadratic cuts to form lower bounds, telling us when exploring a feature further is not worthwhile, (2) the use of the exponential loss, which has an analytical form, obviating the  manipulations needed for logistic loss, (3) the use of a priority queue with a useful ordering function. 

\section*{Code Availability}
Implementations of the fast sparse classification method discussed in this paper are available at \url{https://github.com/jiachangliu/fastSparse}. 

\section*{Acknowledgements}
We acknowledge support from the U.S$.$ National Institutes of Health under NIDA grant DA054994-01, and the National Science Foundation under grant DGE-2022040. We also acknowledge the support of the Natural Sciences and Engineering Research Council of Canada (NSERC).

%% file: sections/10-supplement.tex
\section{THEOREMS AND PROOFS}\label{app:proofs}

\subsection{Thresholding Is Too Conservative}
The first theorem shows that thresholding is too conservative. Recall that with the support set fixed (i.e., $\lambda_0=0$), the loss can be written as $G(\bw) = \sum_{i=1}^n \log(1+\exp(-y_i (\bx_i^{T} \bw))) + \lambda_2 \lVert \bw\rVert_2^2$.

\textbf{Theorem 4.1} \textit{(Thresholding is too conservative.)}
\textit{Let $\bw^{t}$ be the current solution at iteration $t$, $w^t_j$ be the coefficient for the $j$-th feature, and let $w_j^*$ be the optimal value on the $j$-th coefficient while keeping all other coefficients fixed to their values at time $t$.
Furthermore, let $\bw^{t+1}=\bw^t + \be_j (T(j, \bw^t) - w_j^t)$, where $\be_j$ is a vector with $1$ on the $j$-th component and 0 otherwise and $T(j, \bw^t)$ is the thresholding operation with the support set fixed (i.e., $\lambda_0=0$).
Then we have the following inequalities:}
\begin{align*}
    \nabla_{j}G(\bw^t) \nabla_{j} G(\bw^{t+1}) \geq 0, \\
    (w_j^{t} - w_j^*)(w_j^{t+1} - w_{j}^{*}) \geq 0, \\ 
    \text{ and } G(\bw^t) \geq G(\bw^{t+1}).
\end{align*}

\noindent\textit{Proof.}\\
For notational convenience, let us define two functions:
\begin{align*}
    F(u) &:= G(\bw^t) + (u-w_j^t) \nabla_j G(\bw^t) + \frac{1}{2} L_j (u-w_j^t)^2 \\
    H(u) &:= G(\bw^t - w_j^t \be_j + u \be_j).
\end{align*}
Using the notation above, our thresholding operation can be rewritten as $T(j, \bw) \in \argmin_u F(u)$. This means $w_{j}^{t+1}=T(j, \bw)$ minimizes $F(\cdot)$. After the thresholding operation, we update $\bw$ by $\bw^{t+1} = \bw^t -w_j^t \be_j + w^{t+1}_j \be_j$. Furthermore, we use $w_j^*$ to denote the optimal value that minimizes $H(\cdot)$. Throughout this proof, we assume $\lambda_0=0$ because the support set is fixed.

Using the new notation for $F(u)$ and $H(u)$, we have the following expression for their first and second derivatives:
\begin{equation*}
\begin{aligned}[t]
     F'(u) &= \nabla_j G(\bw^t) + L_j (u-w^t_j), \\
     H'(u) &= \nabla_j G(\bw^t - w^t_j \be_j + u \be_j ), \\
     F'(w_j^t) &= H'(w_j^t),
\end{aligned}
\qquad \qquad
\begin{aligned}[t]
     & F''(u) = L_j \\
     & H''(u) = \nabla_{jj}^2 G(\bw^t - w^t_j \be_j + u \be_j ) \\
     & \quad 0 \leq H''(u) \leq L_j = F''(u).
\end{aligned}
\end{equation*}
To get $F'(w_j^t) = H'(w_j^t)$, we plug in $u=w_j^t$ into the formula for $F'(u)$ and $H'(u)$.
For the last inequalities, we have $H''(u) \geq 0$ because $H(u)$ is a convex function. In addition, we have $H''(u) \leq L_j$ because $L_j$ is the Lipschitz constant for $H'(u)$ so that $| H'(u+d) - H'(u) | \leq L_j | d |$ and $| H''(u) | = \lim_{d \to 0} | \frac{H'(u+d) - H'(u)}{d} | \leq L_j$.

Note that $F(u)$ is a quadratic upper bound of $H(u)$. First we have that $F-H$ is a convex function because the second derivative of $F-H$ is greater than or equal to 0.
Second, the first derivative of $F-H$ at $w_j^t$ is 0. Third, $F-H$ at $w_j^t$ is also 0. These three things mean that $F(u)-H(u) \geq 0$ for any $u\in \mathbb{R}$. Therefore, $F(u)$ is a quadratic upper bound of $H(u)$.

We want to show
\begin{align*}
    \nabla_{j}G(\bw^t) \nabla_{j} G(\bw^{t+1}) \geq 0, \\
    (w_j^{t} - w_j^*)(w_j^{t+1} - w_{j}^{*}) \geq 0, \\ 
    \text{ and } G(\bw^t) \geq G(\bw^{t+1}).
\end{align*}

Using the new notation, it is equivalent for us to show
\begin{align}
    H'(w^t_j) H'(w^{t+1}_j) \geq 0, \label{eq:threshold1}\\
    (w_j^{t} - w_j^*)(w_j^{t+1} - w_{j}^{*}) \geq 0, \label{eq:threshold2}\\ 
    \text{ and } H(w^t_j) \geq H(w^{t+1}_j). \label{eq:threshold3}
\end{align}

To show the inequalities above, we discuss three cases: Case 1) $w_j^t < w_j^*$, Case 2) $w_j^t > w_j^*$, and Case 3) $w_j^t = w_j^*$.

\textbf{Case 1}: $w_j^t < w_j^*$

If $w_j^t < w_j^*$, we have $H'(w^t_j) < 0$. This is true because
\begin{align*}
    H'(w^t_j) &= H'(w_j^*) + \int_{w_j^*}^{w_j^t} H''(u) du \\
    &= 0 + \int_{w_j^*}^{w_j^t} H''(u) du \\
    &= - \int^{w_j^*}_{w_j^t} H''(u) du < 0.
\end{align*}
The last inequality holds because $H''(u) \geq 0$ and $H''(u) > 0$ for some nonzero measurable set in $[w_j^t, w_j^*]$.

Now because $w_j^{t+1}$ minimizes $F(\cdot)$, we have $F'(w_j^{t+1}) = 0$. Using the relationship between $F(\cdot)$ and $H(\cdot)$, we have
\begin{align*}
    0 &= F'(w_j^{t+1}) = F'(w_j^t) + \int_{w_j^t}^{w_j^{t+1}} F''(u)du \\
    &= H'(w_j^t) + \int_{w_j^t}^{w_j^{t+1}} F''(u)du \\
    &\geq H'(w_j^t) + \int_{w_j^t}^{w_j^{t+1}} H''(u)du \\
    &= H'(w_j^{t+1}). 
\end{align*}
Therefore, we have $H'(w_j^{t+1}) \leq 0$. Since $H'(w_j^t) < 0$, we have $H'(w_j^t) H'(w_j^{t+1}) \geq 0$, proving \eqref{eq:threshold1} for Case 1.

Let us prove \eqref{eq:threshold2} for Case 1.
For the sake of contradiction, suppose $w_j^{t+1} > w_j^*$, we have $H'(w_j^{t+1}) > 0$ because
\begin{align*}
    H'(w^{t+1}_j) &= H'(w_j^*) + \int_{w_j^*}^{w_j^{t+1}} H''(u) du \\
    &= 0 + \int_{w_j^*}^{w_j^{t+1}} H''(u) du \\
    &= \int^{w_j^{t+1}}_{w_j^*} H''(u) du > 0.
\end{align*}
This implies $H'(w_j^t)H'(w_j^{t+1})< 0$, contradicting the proof of \eqref{eq:threshold1} for Case 1 above. Thus, we have $w_j^{t+1} \leq w_j^*$ and $(w_j^{t} - w_j^*)(w_j^{t+1} - w_j^*) \geq 0$, proving \eqref{eq:threshold2} for Case 1.

Lastly, because of the relationship between $F(\cdot)$ and $H(\cdot)$, we have
\begin{align*}
    H(w_j^{t+1}) \leq F(w_j^{t+1}) = \min_u F(u) \leq F(w_j^t) = H(w_j^t).
\end{align*}
This proves our third inequality \eqref{eq:threshold3} for Case 1.

\textbf{Case 2}: $w_j^t > w_j^*$

If $w_j^t > w_j^*$, we have $H'(w^t_j) > 0$. The procedure to show this is very similar to what we have shown in Case 1, so we omit it here.

Now because $w_j^{t+1}$ minimizes $F(\cdot)$, we have $F'(w_j^{t+1}) = 0$. Using the relationship between $F(\cdot)$ and $H(\cdot)$, we have
\begin{align*}
    0 &= F'(w_j^{t+1}) = F'(w_j^t) + \int_{w_j^t}^{w_j^{t+1}} F''(u)du \\
    &= H'(w_j^t) + \int_{w_j^t}^{w_j^{t+1}} F''(u)du \\
    &\leq H'(w_j^t) + \int_{w_j^t}^{w_j^{t+1}} H''(u)du \\
    &= H'(w_j^{t+1}).
\end{align*}
The third line holds true because $w_j^{t+1} < w_j^t$. Therefore, we have $H'(w_j^{t+1}) \geq 0$. Since $H'(w_j^t) > 0$, we have $H'(w_j^t) H'(w_j^{t+1}) \geq 0$, proving the inequality~\eqref{eq:threshold1} under Case 2.

We proceed to prove \eqref{eq:threshold2} for Case 2. Now for the sake of contradiction, suppose $w_j^{t+1} < w_j^*$, we have $H'(w_j^{t+1}) < 0$. Again, the procedure to show this is very similar to what we have shown in Case 1, so we omit it here.
This reasoning implies $H'(w_j^t)H'(w_j^{t+1})< 0$, contradicting the proof of \eqref{eq:threshold1} under Case 2 above. Thus, we have $w_j^{t+1} \geq w_j^*$ and $(w_j^{t} - w_j^*)(w_j^{t+1} - w_j^*) \geq 0$, proving \eqref{eq:threshold2} for Case 2.

The procedure to prove \eqref{eq:threshold3} under Case 2 identical to \eqref{eq:threshold3} under Case 1, so we omit the proof here.

\textbf{Case 3}: $w_j^t = w_j^*$.

In this special case, $w_j^{t+1} = w_j^t$. Thus, the three inequalities \eqref{eq:threshold1}, ~\eqref{eq:threshold2}, and \eqref{eq:threshold3} hold trivially. This completes the proof for Theorem 4.1.

To the best of our knowledge, we are the first to show that $w_j^{t+1}$ and $w^t_j$ stay on the same side of $w_j^*$. This important point is the motivation behind our use of cutting planes and the development of our quadratic lower bound.

\subsection{Lower Bound via Cutting Planes}
\textbf{Theorem 4.2}
\textit{Suppose $f(x)$ is convex and differentiable on domain $\mathbb{R}$.
Let $\alpha_1$ and $\alpha_2$ be slopes of tangent lines of $f(x)$ at locations $x_1$ and $x_2$.
If $\alpha_1 \alpha_2 \leq 0$, there is a lower bound on the optimal value $f(x^*)$:}
\begin{align*}
    f(x^*) \geq \frac{\alpha_1 f(x_2) - \alpha_2 f(x_1) + \alpha_1 \alpha_2 (x_1 - x_2)}{\alpha_1 - \alpha_2}.
\end{align*}

\noindent\textit{Proof.}\\
Because of the convexity of $f(x)$, we have
\begin{align*}
    f(x) \geq f(x_1) + \alpha_1 (x-x_1) \quad \text{ where } \alpha_1 = f'(x_1)  \\
    f(x) \geq f(x_2) + \alpha_2 (x-x_2) \quad \text{ where } \alpha_2 = f'(x_2). 
\end{align*}

Notice that the function $f(x)$ sits above two lines $y=f(x_1) + \alpha_1 (x-x_1)$ and $y=f(x_2) + \alpha_2 (x-x_2)$.

Equating these two lines to find the intersection point $\hat{x}$, we have
\begin{align}
    & f(x_1) + \alpha_1 (\hat{x}-x_1) = f(x_2) + \alpha_2 (\hat{x}-x_2) \label{eq:linefirst}\\
    \Rightarrow & \hat{x} = \frac{f(x_2)-f(x_1)+\alpha_1 x_1 - \alpha_2 x_2}{\alpha_1 - \alpha_2}.\nonumber
\end{align}
To find the intersection value $\hat{y}$, we plug in $\hat{x}$ into the left side of \eqref{eq:linefirst} and get
\begin{align*}
    \hat{y} &= f(x_1) + \alpha_1 (\hat{x} - x_1) \\
    &= \frac{\alpha_1 f(x_2) - \alpha_2 f(x_1) + \alpha_1 \alpha_2 (x_1 - x_2)}{\alpha_1 - \alpha_2}.
\end{align*}
Since the function $f(x)$ sits above the two lines and therefore above the intersection value $\hat{y}$, we have
\begin{align*}
    f(x^*) \geq \frac{\alpha_1 f(x_2) - \alpha_2 f(x_1) + \alpha_1 \alpha_2 (x_1 - x_2)}{\alpha_1 - \alpha_2}.
\end{align*}
This completes the proof for Theorem 4.2.

\subsection{Lower Bound via Quadratic Cuts}
\textbf{Theorem 4.3} \textit{Suppose $f(x) = g(x)+\lambda_2 x^2$, where $g(x)$ is a convex and differentiable function. Then $f(x)$ is strongly convex.
Let $\alpha_1$ be the slope of the tangent line to $f(x)$ at location $x_1$. Then, there is a lower bound on the optimal value $f(x^*)$:
\begin{align}
    \label{eq:quadratic_cut_1}
    f(x^*) \geq \mathcal{L}_{\textrm{low}}:=f(x_1) - \frac{\alpha_1^2}{4 \lambda_2}.
\end{align}
Let $\alpha_2$ be the slope of the tangent line to $f(x)$ at another location $x_2$.
If $\alpha_1 \alpha_2 \leq 0$, a lower bound on the optimal value $f(x^*)$ is as follows:
\begin{equation}
    \label{eq:quadratic_cut_2}
    f(x^*) \geq \mathcal{L}_{\textrm{low}}:=f(\hat{x}) + \alpha_1 (\hat{x} - x_1) + \lambda_2 (\hat{x} - x_1)^2,
\end{equation}
where
\[
     \hat{x} = \frac{-f(x_1) + f(x_2) + \alpha_1 x_1 - \alpha_2 x_2 - \lambda_2 (x_1^2 - x_2^2) }{\alpha_1 - \alpha_2 - 2\lambda_2 (x_1 - x_2)}. \nonumber
\]}

\noindent\textit{Proof.}\\
Given a convex function $g(x)$, we first show that $f(x) = g(x)+ \lambda_2 x^2$ is a strongly convex function before proving the two bounds in Theorem 4.3.

To show that $f(x)$ is a strongly convex function, it is sufficient to show
\begin{align}\label{eq:convforall}
    f(y) \geq f(x) + f'(x) (y-x) + \lambda_2 (y-x)^2
\end{align}
for any $x, y\in \mathbb{R}$.

Because $g(x)$ is convex, we have
\begin{align*}
    g(y) \geq g(x) + g'(x)(y-x).
\end{align*}
Adding $\lambda_2 y^2$ to both sides, we have
\begin{align*}
    g(y) + \lambda_2 y^2 \geq g(x) + g'(x) (y-x) + \lambda_2 y^2.
\end{align*}
The LHS is $f(y)$. The RHS can be rewritten as
\begin{eqnarray}
    \lefteqn{g(x) + g'(x)(y-x) + \lambda_2y^2} \nonumber\\
    &=& g(x) + \lambda_2 x^2 + (g'(x)+2\lambda_2 x) (y-x) + \lambda_2 y^2 - \lambda_2 x^2 - 2\lambda_2 x(y-x) \nonumber\\
    &=& f(x) + f'(x) (y-x) + \lambda_2 (x-y)^2. \nonumber 
\end{eqnarray}
Therefore, $f(x)$ is a strongly convex function.

Because $f(\cdot)$ is strongly convex, where \eqref{eq:convforall} holds for any $x$ and $y$, given a point $x_1$ with $\alpha_1:=f'(x_1)$, then our strongly convex function $f(\cdot)$ at any point $x$ is bounded by
\begin{align*}
    f(x) \geq f(x_1) + \alpha_1 (x - x_1) + \lambda_2 (x - x_1)^2.
\end{align*}
The RHS is a quadratic function of $x$, with the minimum value achieved at $f(x_1) - \frac{\alpha_1^2}{4\lambda_2}$, so we have
\begin{align*}
    f(x) \geq f(x_1) - \frac{\alpha_1^2}{4\lambda_2}.
\end{align*}
Since the above inequality works for any $x\in \mathbb{R}$, it also works for the optimal value $x^*$:
\begin{align*}
    f(x^*) \geq f(x_1) - \frac{\alpha_1^2}{4\lambda_2}.
\end{align*}
Therefore, we have proved \eqref{eq:quadratic_cut_1}.

Suppose we are given another point $x_2$ with $\alpha_2:=f'(x_2)$, then $f(x)$ sits above two quadratic equations:
\begin{align*}
    f(x) &\geq f(x_1) + \alpha_1 (x - x_1) + \lambda_2 (x - x_1)^2 \\
    f(x) &\geq f(x_2) + \alpha_2 (x - x_2) + \lambda_2 (x - x_2)^2.
\end{align*}
Equating these two quadratic equations to find the intersection point $\hat{x}$, we have
\begin{align*}
    &f(x_1) + \alpha_1 (\hat{x} - x_1) + \lambda_2 (\hat{x} - x_1)^2 = f(x_2) + \alpha_2 (\hat{x} - x_2) + \lambda_2 (\hat{x} - x_2)^2\\
    \Rightarrow & \hat{x} = \frac{-f(x_1)+f(x_2) + \alpha_1 x_1 -\alpha_2 x_2  + \lambda_2 (x_2^2 - x_1^2)}{\alpha_1 - \alpha_2 -2\lambda_2 (x_1 - x_2)}.
\end{align*}
Plugging in the intersection point $\hat{x}$, we can get the intersection value $\hat{y}$, which is a lower bound of $f(x^*)$
\begin{align*}
    f(x^*) \geq \hat{y} = f(x_1) + \alpha_1 (\hat{x} - x_1) + \lambda_2 (\hat{x} - x_1)^2.
\end{align*}
This completes the proof for \eqref{eq:quadratic_cut_2}.

\subsection{Derivation for the Exponential Loss}\label{app:proofs_exp}
The exponential loss function is defined as $H (\bw) = \sum_{i=1}^n e^{-y_i f(\mathbf{x}_i)}$, where $f(\bx_i)=\bw^T\bx_i$. 
Since $\bx_i$ is a binary vector, s.t. $x_{ij} \in \{-1,1\}$ and $y_i \in \{-1,1\}$, let $\bz_i = y_i\bx_i$ and $\bz_i \in \{-1,1\}^p$. After $t$ iterations, the exponential loss function can be written as: $$H(\bw^t) = \sum_{i=1}^n e^{-y_i (\sum_{j=1}^p w_j^t x_{ij})} = \sum_{i=1}^n e^{-(\bw^t)^T\bz_i}
.$$

We will perform a linesearch, where we optimize coefficient $j$ at iteration $t$. This linesearch optimization problem for coordinate $j$ is $w_j^{t+1}\in\argmin_w H( [w_1^t,...,w_{j-1}^t,w,w_{j+1}^t,...])+\lambda_0\|[w_1^t,...,w_{j-1}^t,w,w_{j+1}^t,...]\|_0$.

\textbf{Theorem 5.1} 
\textit{Let $\bw^t$ be the coefficient vector at iteration $t$, $H^t:=H(\bw^t)$ and $\lambda_0$ be the regularization constant for the $\ell_0$ penalty. For the $j$-th coordinate, we update the coefficient according to:}

\textit{(1) Suppose $w_j^t \!= \!0$. Let $d_- \!=\! \sum_{i: z_{ij}=-1}\!c_i / \!\sum_{i = 1}^n \! c_i$, where $c_i = e^{-(\bw^t)^T \bz_i}$. Then, if $d_-$ is within the interval:}
\[
\left[\frac{1}{2}\!-\!\frac{1}{2H^t}\sqrt{\lambda_0(2H^t\!-\!\lambda_0)},\frac{1}{2}\!+\!\frac{1}{2H^t}\sqrt{\lambda_0(2H^t\!-\!\lambda_0)}\right],
\]
\textit{then set $w_{j}^{t+1}$ to 0. Otherwise set $w_{j}^{t+1} = \frac{1}{2}\ln \frac{1-d_-}{d_-}$.}

\textit{(2) Suppose $w_j^t \!\neq\! 0$.  Let $D_- \!=\! \sum_{i: z_{ij}=-1}\!c_i / \!\sum_{i = 1}^n \! c_i$, where $c_i = e^{-(\bw^t - w^t_j \be_j)^T z_i}$. Let $H^t_{\neg j} = H(\bw^t-w^t_j \be_j)$. 
Then, if $D_-$ is within the interval:}
\[
\left[\frac{1}{2}\!-\!\frac{1}{2H^t_{\neg j}}\sqrt{\lambda_0(2H^t_{\neg j}\!-\!\lambda_0)},\frac{1}{2}\!+\!\frac{1}{2H^t_{\neg j}}\sqrt{\lambda_0(2H^t_{\neg_j}\!-\!\lambda_0)}\right],
\]
\textit{then set $w_j^{t+1}$ to 0. Otherwise, set $w_j^{t+1}=\frac{1}{2}\ln \frac{1-D_-}{D_-}$.}

While these expressions may first appear difficult to calculate, they are not. Like AdaBoost, we make multiplicative updates to the loss at each iteration. Thus, since $H^t$ is easy to calculate, $H^t_{\neg j}$ is also easy to calculate (requiring only a multiplication), and the rest is simple mathematical operations. 

Intuitively, using AdaBoost's terminology, the bound states that if the weak learning algorithm produces a stronger weak classifier at that iteration (a classifier whose error rate is away from 1/2), we would keep it. Otherwise, we would not; we would rather set its coefficient to 0. In some sense, this result is reminiscent of iterative thresholding \citep{daubechies2004iterative}.

\noindent\textit{Proof.}\\
\noindent\textbf{Case 1}: 
Suppose at iteration $t$, $w_j^t = 0$ and in the next iteration $t+1$, we evaluate placing feature $j$ into the model, i.e., set $w_j^{t+1} \neq 0$. Then, the decrease in loss should be larger than $\lambda_0$, otherwise $w_j^{t+1}=0$. 
Suppose we want to add feature $j$ into the model, the loss function is $$H^{t+1} = \sum_{i=1}^n e^{-(\bw^t)^T \bz_i - y_i w_j x_{ij}} = \sum_{i=1}^n e^{-(\bw^t)^T \bz_i - w_j z_{ij}}.$$ We can get an analytical solution for $w_j$ by solving $\frac{\partial H^{t+1}}{\partial w_j} =0$, which is the same as AdaBoost's update step.
\begin{equation}\label{eq:derivative}
    \begin{aligned}
    0=\frac{\partial H^{t+1}}{\partial w_j}\Big|_{w_j^*}
    =& \sum_{i=1}^n -z_{ij} e^{-(\bw^t)^T \bz_i} e^{-w_j z_{ij}}\Big|_{w_j^*}\\
    =& \sum_{i: z_{ij} = 1} -e^{-(\bw^t)^T \bz_i} e^{-w_j} \Big|_{w_j^*}+ 
    \sum_{i: z_{ij}=-1} e^{-(\bw^t)^T \bz_i} e^{w_j}\Big|_{w_j^*}. 
    \end{aligned}
\end{equation}
Multiplying by a normalization constant $$C = \sum_{i=1}^n c_i = \sum_{i=1}^n e^{-(\bw^t)^T \bz_i},$$ and defining $$d_+ = \frac{\sum_{i: z_{ij}=1} e^{-(\bw^t)^T \bz_i}}{C} \quad \textrm{and} \quad d_- = \frac{\sum_{i: z_{ij}=-1} e^{-(\bw^t)^T \bz_i}}{C},$$ 
Equation \eqref{eq:derivative} becomes
\[0=-d_+e^{-w_j^*}+d_-e^{w_j^*}.\]
Solving this yields:
\[w^*_j =\frac{1}{2}\ln \frac{d_+}{d_-}.\]

Recalling that $d_+=1-d_-$, the lowest possible loss after adding in feature $j$ is thus: 
\begin{equation}\label{eq:wkbest}
H^{t+1} = \mathcal{L}^t \cdot \left((1-d_-)\left(\frac{1-d_-}{d_-}\right)^{-1/2} + d_-\left(\frac{1-d_-}{d_-}\right)^{1/2}\right) = H^t \cdot 2 \left( (1-d_-)d_- \right)^{1/2}.
\end{equation}
We have now derived the best possible value for $w_j$ if it were nonzero. However, our objective suffers a penalty of $\lambda_0$ from the regularization term whenever $w_j$ is nonzero. Thus, we need to compare the objective with $w_j=0$ to the regularized objective with \eqref{eq:wkbest} as the loss term. If the difference is less than $\lambda_0$, it would benefit the objective to set coefficient $j$ to 0 at the next iteration. The condition for setting $w_j$ to 0 is:
$$H^{t} - H^{t+1} = H^t -  H^{t} \cdot 2 \left( (1-d_-)d_- \right)^{1/2} \leq \lambda_0.$$
\begin{equation}\label{eq:quadratic_eq}
    \begin{aligned}
     \frac{H^t - \lambda_0}{2H^t} &\leq ((1-d_-)d_-)^{1/2} \\ 
     \left(\frac{H^t - \lambda_0}{2H^t}\right)^2 &\leq (1-d_-)d_-\\
     d_-^2 - d_- + \left(\frac{H^t - \lambda_0}{2H^t}\right)^2 &\leq 0.
    \end{aligned}
\end{equation}
This is a quadratic equation, permitting solutions in $d_- \in \left(\frac{1}{2} - \frac{1}{2H^t} \sqrt{\lambda_0(2H^t - \lambda_0)}, \frac{1}{2} + \frac{1}{2H^t} \sqrt{\lambda_0(2H^t - \lambda_0)}\right)$. 

Therefore, if $d_- \in \left(\frac{1}{2} - \frac{1}{2H^t} \sqrt{\lambda_0(2H^t - \lambda_0)}, \frac{1}{2} + \frac{1}{2H^t} \sqrt{\lambda_0(2H^t - \lambda_0)}\right)$,
then set $w_j^{t+1}$ to 0. Otherwise, $w_j^{t+1} = \frac{1}{2}\ln \frac{1-d_-}{d_-}$. 

\noindent\textbf{Case 2:}
Suppose at iteration $t$, $w_j^t \neq 0$, and in the next iteration $t+1$, we evaluate updating $w_j^t$. Then the decrease in loss should be larger than $H^t - H^t_{\neg j} + \lambda_0$, 
otherwise, $w_j^{t+1 = 0}$. 
Suppose we want to update $w_j$ at iteration $t+1$, the loss function is  
$$H^{t+1} = \sum_{i=1}^n e^{-(\bw^t-w_j^t\be_j)^T \bz_i- w_j z_{ij}}\;\;.$$

Similar to the derivation for Case 1, we can get an analytical solution for $w_j$ by solving $\frac{\partial H^{t+1}}{\partial w_j} = 0$. 

\begin{equation}\label{eq:derivative_case2}
    \begin{aligned}
    0 = \frac{\partial H^{t+1}}{\partial w_j} \big|_{w_j^*}
    =& \sum_{i=1}^n -z_{ij} e^{-(\bw^t-w_j^t\be_j)^T \bz_i} e^{(-w_j z_{ij})} \big|_{w_j^*}\\
    =& \sum_{i: z_{ij} = 1} -e^{-(\bw^t-w_j^t\be_j)^T \bz_i} e^{-w_j} \big|_{w_j^*} + \\
    &\sum_{i: z_{ij}=-1} e^{-(\bw^t-w_j^t\be_j)^T \bz_i} e^{w_j} \big|_{w_j^*}\;\;.
    \end{aligned}
\end{equation}
Similarly, multiplying by a normalization constant $C$, and defining $D_+ = \sum_{i: z_{ij} = 1} e^{-(\bw^t-w_j^t\be_j)^T \bz_i}/C$ and $D_-=\sum_{i: z_{ij} = -1} e^{-(\bw^t-w_j^t\be_j)^T \bz_i}/C$.

Then Equation \ref{eq:derivative_case2} becomes 
$$0 = -D_+ e^{-w_j^*} + D_-e^{w_j^*}.$$
Solving this yields:
$$w_j^* = \frac{1}{2}\ln \frac{D_+}{D_-}.$$

The lowest possible loss after updating the coefficient of feature $j$ is 
\begin{equation}\label{eq:wkbest_case2}
    H^{t+1} = H^t_{\neg j} \cdot \left(D_+\left(\frac{D_+}{D_-}\right)^{-1/2} + D_-\left(\frac{D_+}{D_-}\right)^{1/2}\right) = H^t_{\neg j} \cdot 2((1-D_-)D_-)^{1/2}.
\end{equation}

Similarly to Case 1, we need to compare the objective with $w_j=0$ to the regularized objective with \eqref{eq:wkbest_case2} as the loss term. If the difference is less than $\lambda_0$, it would benefit the objective to set coefficient $j$ to 0 at the next iteration. The condition for setting $w_j$ to 0 is:
$$H^{t}_{\neg j} - H^{t+1} = H^t_{\neg j} -  H^{t}_{\neg j} \cdot 2 \left( (1-D_-)D_- \right)^{1/2} \leq \lambda_0.$$ 
Using the same derivation as in Equation \eqref{eq:quadratic_eq}, 
the solution is in $$D_- \in \left(\frac{1}{2} \!-\! \frac{1}{2H^t_{\neg j}} \sqrt{\lambda_0(2 H^t_{\neg j}\! -\! \lambda_0)}, \frac{1}{2} + \frac{1}{2H^t_{\neg j}} \sqrt{\lambda_0(2 H^t_{\neg j} - \lambda_0)}\right).$$ 

Therefore, if $D_- \in \left(\frac{1}{2} \!-\! \frac{1}{2H^t_{\neg j}} \sqrt{\lambda_0(2 H^t_{\neg j}\! -\! \lambda_0)}, \frac{1}{2} + \frac{1}{2H^t_{\neg j}} \sqrt{\lambda_0(2 H^t_{\neg j} - \lambda_0)}\right)$, then set $w_j^{t+1}$ to 0. Otherwise, $w_j^{t+1} = \frac{1}{2}\ln \frac{1-D_-}{D_-}$. 

\section{PSEUDOCODE}\label{app:pseudocode}

We begin with the presentation of our high-level Algorithm \ref{alg:overall_alg} and then elaborate on the novel steps in the following lower-level algorithms.

Shortly, we discuss how \textrm{TryDeleteOrSwap}$(\bw, j, S^c)$ is implemented in detail. After that, we discuss its subroutine algorithms \textrm{TryAddLinCut}$(\bw', j', \mathcal{L}_{best})$ and \textrm{TryAddQuad}$(\bw', j', \mathcal{L}_{best})$, as well as their subroutine algorithm \textrm{FindNewCoefficient}$(\bw', j')$.
For algorithms \textrm{TryAddLinCut}$(\bw', j', \mathcal{L}_{best})$ and \textrm{TryAddQuad}$(\bw', j', \mathcal{L}_{best})$, we use $f(x) = G(\bw'+\be_{j'}x)$ for notational convenience (assuming $w'_{j'} = 0$; if it is not, notation can be adjusted appropriately).
Also for notational convenience, we use one lower bound from classical cutting planes and two lower bounds from quadratic cuts:
\begin{align*}
    \text{LinCut}(a, b, f(\cdot)) &= \frac{f'(a) f(b) - f'(b) f(a) + f'(a) f'(b) (a - b)}{f'(a) - f'(b)}\\
    \text{QuadCut1}(a, f(\cdot)) &= f(a) - \frac{f'(a)^2}{4\lambda_2}\\
    \text{QuadCut2}(a, b, f(\cdot)) &= f(a) + f'(a) (\hat{x} - a) + \lambda_2 (\hat{x} - a)^2 \\
    \text{with } \hat{x} &= \frac{-f(a)+f(b) + f'(a) a - f'(b) b  + \lambda_2 (b^2 - a^2)}{f'(a) - f'(b) -2\lambda_2 (a - b)}.
\end{align*}

\begin{algorithm}[!h]
\caption{General Algorithm for Swapping Features}
\textbf{Input:} coefficients $\bw$ from a warm start algorithm, $\bc = \mathbf{0}$ is a vector of size $p$ where each $c_j$ for $j < p$ indicates the number of times we failed to find a feature to swap with $j$.\\
\textbf{Output:} updated coefficients $\bw$ that is a swap 1-OPT solution.\hfill
\begin{algorithmic}[1]
    \WHILE{True}
        \STATE Update support $S=\{j|w_j \neq 0\}$. \label{alg:overall_alg_restart}
        \STATE $\Pi(S)$ = Sort($S$) according to $c_j$ for $j\in S$. \hfill\#\textit{Sort support in ascending order of the no$.$ of failed swaps} 
        \FOR{$j$ in $\Pi(S)$}
            \STATE $\bw' = \text{TryDeleteOrSwap}(\bw, j, S^c)$ \hfill\#\textit{$S^c$ is the complement of $S$}
            \IF {$\bw' \neq \bw$}
                \STATE Let $\bw = \bw'$. \hfill\#\textit{Swap was successful}
                \STATE Go to line 2.
            \ELSE
                \STATE $c_j = c_j + 1$. \hfill\#\textit{No better feature can replace feature $j$}
            \ENDIF
        \ENDFOR
        \STATE Return $\bw$. \hfill\#\textit{No single feature can be replaced with better features}
    \ENDWHILE
    
\end{algorithmic}
 \label{alg:overall_alg}
\end{algorithm}

\newpage
\begin{algorithm}[ht]
\caption{TryDeleteOrSwap($\bw, j, S^c$)}
\textbf{Input:} coefficients $\bw$, feature index $j$ with $w_j \neq 0$, set of feature indices $S^c=\{j'|w_{j'}=0\}$.\\
\textbf{Output:} updated coefficients $\bw'$ with feature $j$ possibly deleted or swapped with feature $j'\in S^c$.\\
\begin{algorithmic}[1]
    \STATE Calculate the best current loss $\mathcal{L}_{best} = G(\bw)$.
    \STATE Let $\bw' = \bw$ and then set $w'_j = 0$. \hfill\#\textit{Drop feature $j$ from the support}
    \IF{$G(\bw') \leq \mathcal{L}_{best}$}
        \STATE Update $\bw'$ with support restricted to $S\setminus \{j\}$. \hfill\#\textit{Dropping feature $j$ leads to smaller loss}
        \STATE Return $\bw'$.
    \ENDIF
    \STATE Calculate $|\nabla_{S^c} G(\bw^t)|$ on $S^c$.
    \STATE $\Pi'$ = feature indices in $S^c$ sorted in descending order of $|\nabla_{S^c} G(\bw')|$. \hfill\#\textit{Order features to possibly add in}
    \FOR {$j' \in \Pi'$} \label{alg:swap_firstLoop_start}
        \STATE Let $\bw' = \text{TryAddQuad}(\bw', j', \mathcal{L}_{best})$ if $(\lambda_2>0)$. \hfill\#\textit{or $\bw' = \text{TryAddLinCut}(\bw', j', \mathcal{L}_{best})$ if $(\lambda_2=0)$}
        \IF {$\bw'$ has changed}
            \STATE Update full vector $\bw'$ with support restricted to $S \cup \{j'\} \setminus \{j\} $. \hfill\#\textit{Swapping $j$ with $j'$ decreases loss}
            \STATE Return $\bw'$.
        \ENDIF
    \ENDFOR \label{alg:swap_firstLoop_end}
    \STATE Return $\bw$.\hfill\#\textit{Since there were no better features, return original $\bw$}
\end{algorithmic}
 \label{alg:dynamic_ordering}
\end{algorithm}
 
\begin{algorithm}[!h]
\caption{TryAddLinCut($\bw', j', \mathcal{L}_{best}$)}
\textbf{Input:} coefficients $\bw'$, feature index $j'$, and current best loss $\mathcal{L}_{best}$.\\
\textbf{Output:} updated coefficients $\bw'$.
\begin{algorithmic}[1]
     \STATE Let $a=T(j', \bw')$, $b=2 T(j', \bw')$ \hfill\#\textit{Take 2X distance suggested by thresholding operation Eq \eqref{formula:thresholding}}
    \IF {$f'(0) f'(b) < 0$}
        \STATE Let $c=(a+b)/2$ \label{alg:lincut_firstLoop_start} \hfill\#\textit{Binary search} 
        \IF {$f'(0) f'(c) < 0$} 
            \STATE Let $b=c$
        \ELSE
            \STATE Let $a=c$
        \ENDIF \hfill\#\textit{a and b are on opposite sides of $w_{j'}^*$}
        \STATE Get $\mathcal{L}_{low} = \text{LinCut} (a, b, f(\cdot))$
        \IF {$\mathcal{L}_{low} \geq \mathcal{L}_{best}$}
            \STATE Return $\bw'$ \hfill\#\textit{Stop considering feature $j'$ and exit early}
        \ENDIF
        \STATE Let $\hat{w}_{j'}=\text{FindNewCoefficient}(\bw', j')$. \label{alg:lincut_secondLoop_start}
        \IF {$f(\hat{w}_{j'}) < \mathcal{L}_{best}$}
            \STATE Return $\bw'$ with $w'_{j'} = \hat{w}_{j'}$ \hfill\#\textit{Swap feature $j$ with feature $j'$}
        \ENDIF
        \STATE Return $\bw'$ \hfill\#\textit{Eliminate considering feature $j'$}  \label{alg:lincut_secondLoop_end}
    \ENDIF
    \STATE Let $a=2T(j', \bw')$, $b=3T(j', \bw')$ \hfill\#\textit{Take 3X distance suggested by thresholding operation}
    \IF {$f'(0) f'(b) < 0$}
        \STATE Go to line 9. 
    \ELSE 
        \STATE Go to line 13. 
        \hfill\#\textit{Minimum is far from starting point. Swap the feature to see if there's improvement.}
    \ENDIF
\end{algorithmic}
 \label{alg:linear_cut}
\end{algorithm}
 
\newpage
\begin{algorithm}[!ht]
\caption{TryAddQuad($\bw', j', \mathcal{L}_{best}$)}
\textbf{Input:} coefficients $\bw'$, feature index $j'$, and current best loss $\mathcal{L}_{best}$ \\
\textbf{Output:} updated coefficients $\bw'$
\begin{algorithmic}[1]
    \STATE Get $\mathcal{L}_{low} = \text{QuadCut1}(0, f(\cdot))$
    \IF{$\mathcal{L}_{low} \geq \mathcal{L}_{best}$}
        \STATE Return $\bw'$ \hfill\#\textit{Stop considering feature $j'$ and exit early}
    \ENDIF
    \STATE Let $a=T(j', \bw')$, $b=2 T(j', \bw')$ \hfill\#\textit{Take 2X distance suggested by thresholding operation Eq \eqref{formula:thresholding}}
    \IF {$f'(0) f'(b) < 0$}
        \STATE Let $c=(a+b)/2$ \hfill\#\textit{Binary search} \label{alg:quadcut_firstLoop_start}
        \STATE Get $\mathcal{L}_{low}=\text{QuadCut1}(c, f(\cdot))$
        \IF{$\mathcal{L}_{low} \geq \mathcal{L}_{best}$}
            \STATE Return $\bw'$ \hfill\#\textit{Stop considering feature $j'$ and exit early}
        \ENDIF
        \IF {$f'(0) f'(c) < 0$}
            \STATE Let $b=c$
        \ELSE
            \STATE Let $a=c$
        \ENDIF \hfill\#\textit{a and b are on opposite sides of $w_{j'}^*$}
        \STATE Get $\mathcal{L}_{low} = \text{QuadCut2}(a, b, f(\cdot))$
        \IF {$\mathcal{L}_{low} \geq \mathcal{L}_{best}$}
            \STATE Return $\bw'$ \hfill\#\textit{Stop considering feature $j'$ and exit early}
        \ENDIF
        \STATE Let $\hat{w}_{j'}=\text{FindNewCoefficient}(\bw', j')$. \label{alg:quadcut_secondLoop_start}
        \IF {$f(\hat{w}_{j'}) < \mathcal{L}_{best}$}
            \STATE Return $\bw'$ with $w'_{j'} = \hat{w}_{j'}$ \hfill\#\textit{Swap feature $j$ with feature $j'$}
        \ENDIF
        \STATE Return $\bw'$ \hfill\#\textit{Eliminate considering feature $j'$} \label{alg:quadcut_secondLoop_end}
    \ENDIF
    \STATE Let $a=2T(j', \bw')$, $b=3T(j', \bw')$ \hfill\#\textit{Take 3X distance suggested by thresholding operation}
    \STATE Get $\mathcal{L}_{low}=\text{QuadCut1}(a, f(\cdot))$
    \IF{$\mathcal{L}_{low} \geq \mathcal{L}_{best}$}
        \STATE Return $\bw'$ \hfill\#\textit{Stop considering feature $j'$ and exit early}
    \ENDIF
    \IF {$f'(0) f'(b) < 0$}
        \STATE Go to line 17. 
    \ELSE
        \STATE Get $\mathcal{L}_{low}=\text{QuadCut1}(b, f(\cdot))$
        \IF{$\mathcal{L}_{low} \geq \mathcal{L}_{best}$}
            \STATE Return $\bw'$ \hfill\#\textit{Stop considering feature $j'$ and exit early}
        \ENDIF
        \STATE Go to line 21. 
        \hfill\#\textit{Minimum is far from starting point. Swap the feature to see if there's improvement.}
    \ENDIF
\end{algorithmic}
 \label{alg:quadratic_cut}
\end{algorithm}

\begin{algorithm}[h!]
\caption{FindNewCoefficient($\bw', j'$)}
\textbf{Input:} coefficients $\bw$, coordinate $j'$, iteration steps max\_iter=10 (default)\\
\textbf{Output:} updated coefficient $w_{j'}$ for coordinate $j'$
\begin{algorithmic}[1]
    \FOR{t in 1, 2, ..., max\_iter}
        \STATE $\bw' = \bw' - w_{j'} \be_{j'} + T(\bw', j') \be_{j'}$ \hfill\#\textit{Apply the thresholding operation on coordinate $j'$}
    \ENDFOR
    \STATE Return $w'_{j'}$
\end{algorithmic}
 \label{alg:iterative_thresholding}
\end{algorithm}

\newpage
\section{EXPERIMENTAL DETAILS}\label{app:exp_details}

We next present the datasets used in our experiments, our preprocessing steps, and the experimental setup. 

\subsection{Datasets}
We present results using 5 datasets: two synthetic datasets (one in which the features are highly correlated for binary classification and the other in which the features are highly correlated for linear regression), the Fair Isaac (FICO) credit risk dataset \citep{fico} used for the Explainable ML Challenge, two recidivism datasets: COMPAS \citep{LarsonMaKiAn16} and Netherlands \citep{tollenaar2013method}.
We predict whether an individual will default on a loan for the FICO dataset, which individuals are arrested within two years of release on the COMPAS dataset, and whether defendants have any type of charge within four years on the Netherlands dataset. 

\begin{table}[ht]
\centering
\begin{tabular}{l||cc}
\toprule[1.2pt]
Dataset Name & n  & p  \\ \hline
Highly Correlated (classification)            &         800              &        1000                   \\
Highly Correlated (regression)            &         2000              &        2000                   \\
FICO             &       10459      &           1917             \\
COMPAS             &        6907        &  134                      \\
NETHERLANDS     &  20000            & 2024                       \\
\bottomrule[1.2pt]
\end{tabular}
\caption{Datasets and their number of samples (n) and number of features (p).}
\label{tab:dataset_details}
\end{table}

\subsection{Data Generation and Preprocessing}
\textbf{Synthetic Datasets}

\textbf{Binary Classification:} we generate synthetic datasets according to the generation process in L0Learn~\citep{dedieu2020learning}. We first sample the data features $\bx_i \in \mathbb{R}^{p}$ from a multivariate Gaussian distribution $\mathcal{N}(0, \Sigma)$ with mean $0$ and covariance matrix $\Sigma$.
Then, we create the coefficient vector $\bw$ with $k$ nonzero entries, where $w_i = 1$ if $i \textrm{ mod }(p/k) = 0$.
Lastly, we sample the data labels $y_i \in \{-1, +1\}$ from a Bernoulli distribution $P(y_i=1\mid \bx_i) = \frac{1}{1+\exp(-\bw^{T} \bx_i)}$.
In our experiments, we generate $800$ training and  $160$ test samples with feature dimension $p=1000$. The data are highly correlated with $\Sigma_{ij} = 0.9^{|i-j|}$. Additionally, we set the number of true sparsity $k=25$. We generate this setting 5 times with 5 different random seeds (in total we have 5 datasets, each with $(800+160) = 960$ samples).

\textbf{Linear Regression:}  we generate the synthetic dataset according to the generation process in L0Learn~\citep{hazimeh2020fast} as explained in Section 5.3.1. We first sample the data features $\bx_i \in \mathbb{R}^{p}$ from a multivariate Gaussian distribution $\mathcal{N}(0, \Sigma)$ with mean $0$ and covariance matrix $\Sigma$.
Then, we create the coefficient vector $\bw$ with $k$ nonzero entries, where $w_i = 1$ if $i \textrm{ mod } (p/k) = 0$.
Lastly, we sample $y_i = \bx_i^{T} \bw + \epsilon_i$ with $\epsilon_i$ generated from a Gaussian distribution $\mathcal{N}(0, \sigma^2)$. The signal-to-noise ratio (SNR) is defined as $\text{SNR} = \frac{\text{Var}(X\bw)}{\text{Var}(\epsilon)} = \frac{\bw^{T} \Sigma \bw}{\sigma^2}$, where each row of $X$ is $\bx_i$.
In our experiments, we generate $2000$ data samples with feature dimension $p=2000$. The data are highly correlated with $\Sigma_{ij} = 0.9^{|i-j|}$ and $\text{SNR}=5$. Additionally, we set the true sparsity as $k=100$.

\noindent\textbf{Real Datasets}

\textbf{FICO:} We use all continuous features in this dataset. We did not consider missing data values as separate dummy variables.

\textbf{COMPAS:} We selected features \textit{sex, age, juv\_fel\_count, juv\_misd\_count, juv\_other\_count, priors\_count}, and \textit{c\_charge\_degree}
and the label \textit{two\_year\_recid}.

\textbf{NETHERLANDS:} We translated the feature names from Dutch to English and then used features \textit{sex, country of birth, log \# of previous penal cases, 11-20 previous case, and $>$20 previous case, age in years, age at first penal case, offence type}, and the label \textit{recidivism\_in\_4y}.

For FICO and COMPAS, we convert each continuous variable $x_{\cdot,j}$ into a set of highly correlated dummy variables $\tilde{x}_{\cdot,j,\theta} = \bm{1}_{[x_{\cdot,j}\leq \theta]}$, where $\theta$ are all unique values that have appeared in feature column $j$. For NETHERLANDS, we convert continuous variables into a set of dummy variables in the same way except for variables \textit{age in years} (which is real-valued, not integer) and \textit{age at first penal case}. For these two real-valued variables, instead of considering all unique values that have appeared in the feature column, we consider 1000 quantiles.

\subsection{Evaluation Platform}
All experimental results were run on a 2.40GHz 30M Cache (256GB RAM 48 hyperthreaded cores) Dell R620 with 2 Xeon(R) CPU E5-2695 v2. We ran all experiments using 8 cores per task.

\subsection{Software Packages Used}
We list all software packages used in this section. Details about hyperparameter selection are in Appendix \ref{app:exp_results}. 
\begin{itemize}
    \item $\ell_1$ regularized logistic regression: We run $\ell_1$ regularized logistic regression using \textit{glmnet} package \citep{glmnet}. 
    \item Minimax Concave Penalty (MCP):  We run MCP using \textit{ncvreg} package \citep{ncvreg}.  
    \item L0Learn: We run L0Learn using the R implementation from \citep{dedieu2020learning}\footnote{https://github.com/hazimehh/L0Learn}. 
    \item Ours: We build our method based on L0Learn's codebase, so that we could use its preprocessing steps, and pipeline for running the full regularization path of $\lambda_0$ values.
\end{itemize}
There are some other baselines such as GraSP~\citep{bahmani2013greedy} and NHTP~\citep{zhou2021global}. However,  previous work~\citep{dedieu2020learning} has shown that they have a considerable number of false positives on the synthetic dataset and have large support sizes for their solutions, so we omit running these two baselines.

\subsection{Evaluation Metrics}
We use the same evaluation metrics used in \citet{dedieu2020learning}. 
\begin{itemize}
    \item AUC: The area under the ROC curve. 
    \item Accuracy: $1-\frac{\sum_{i=1}^n \mathbb{1}[y_i \neq \hat{y}_i]}{n}$. 
    \item Recovery F1 score: $\frac{2PR}{P+R}$, where $P = |\text{supp}(\hat{\bw}) \cap \text{supp}(\bw^*)| / |\text{supp}(\hat{\bw})|$ is the precision and $R = |\text{supp}(\hat{\bw}) \cap \text{supp}(\bw^*)| / |\text{supp}({\bw}^*)|$ is the recall. $\text{supp}(\cdot)$ stands for the support (indices with nonzero coefficients) of a solution. We can only use recovery F1 score for synthetic datasets since we need to know the support of ${\bw}^*$ to calculate it. 
\end{itemize}

\section{ADDITIONAL EXPERIMENTS}\label{app:exp_results}
We first elaborate on hyperparameters used for different software packages. We then present extra experimental results that were omitted from the main paper due to space constraints. 

\noindent\textbf{Collection and Setup:} we ran the experiments on the two simulated datasets and 3 real datasets: FICO, COMPAS, and Netherlands. For each dataset, we trained the model using varying configurations. On the simulated classification task, we ran on 5 datasets, each generated by a different random seed. On the real datasets, we performed 5-fold cross validation to measure training time, training accuracy, and test accuracy for each fold.

To get the support versus AUC, accuracy, and F1 score curves, for MCP, the sequence of 100 $\lambda$ values was
set to the default values of ncvreg, and we chose the second parameter $\gamma$ by using 10 values between 1.5 and 25, where we show results of $\gamma$ being 1.5 and 25 for each $\lambda$. Curves for other $\gamma$ values are between these extremes. For $\ell_1$ regularized logistic regression, the choice of 100 $\lambda$ values was set to the default
sequence chosen by glmnet. 
For L0Learn, we set the penalty type to ``L0L2'' and $\gamma$ to $\{0.00001, 0.0001, 0.001, 0.01, 0.1, 1, 10\}$. The regularization choices for the $\ell_0$ term were set to the 100 default values of L0Learn.
For our methods, we also set the penalty type and $\gamma$ (which is $\lambda_2$) in the same way as the setting for L0Learn and use the same $\lambda$ values as in the L0Learn algorithm.

In addition, we set 56 pairs of $\lambda$ (resp. $\lambda_0$) and $\gamma$ (resp. $\lambda_2$) values for comparing the run times obtained by our methods and by L0Learn: $\lambda \in \{0.8, 1, 2, 3, 4, 5, 6, 7\}$ and $\gamma \in \{0.00001, 0.0001, 0.001, 0.01, 0.1, 1, 10\}$. 

\subsection{Run Time Savings for Linear Regression}\label{app:linear_regression}
Although our method is designed for classification problems, the proposed dynamic ordering technique can also speed up the local swap process for linear regression. For the full regularization path with 100 different $\lambda_0$ values, the total time difference between local swaps without dynamic ordering and local swaps with dynamic ordering improved computation time by $36\%$ (from 184 seconds to 117 seconds).

\subsection{Run Time Savings from First and Second Methods}\label{app:run_time_savings}

\begin{figure*}[t]
  \centering
  \includegraphics[width=1.0\linewidth]{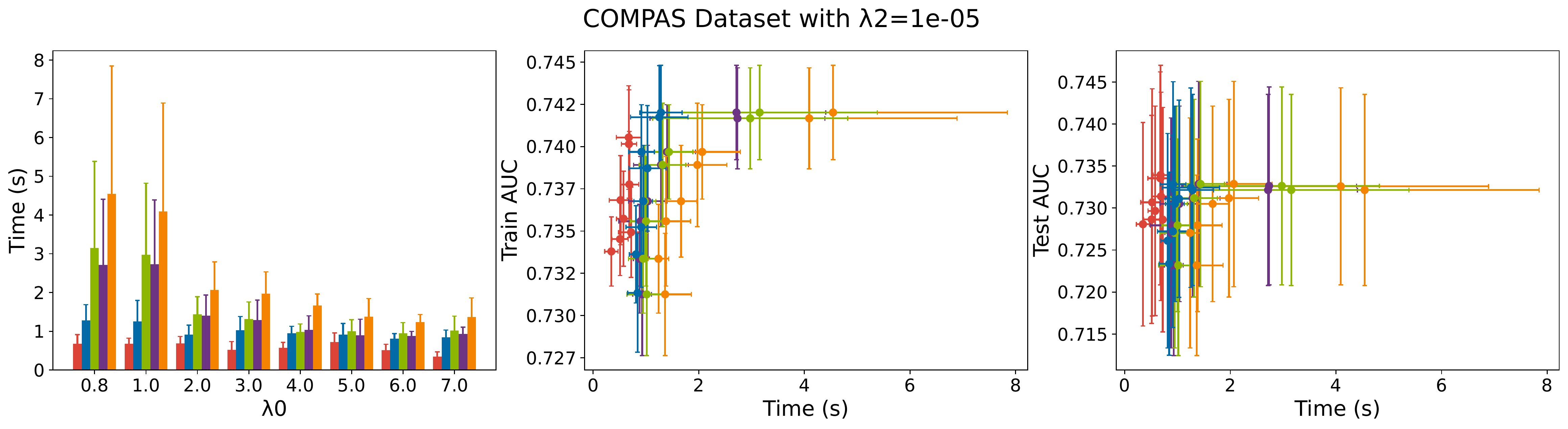}

  \includegraphics[width=1.0\linewidth]{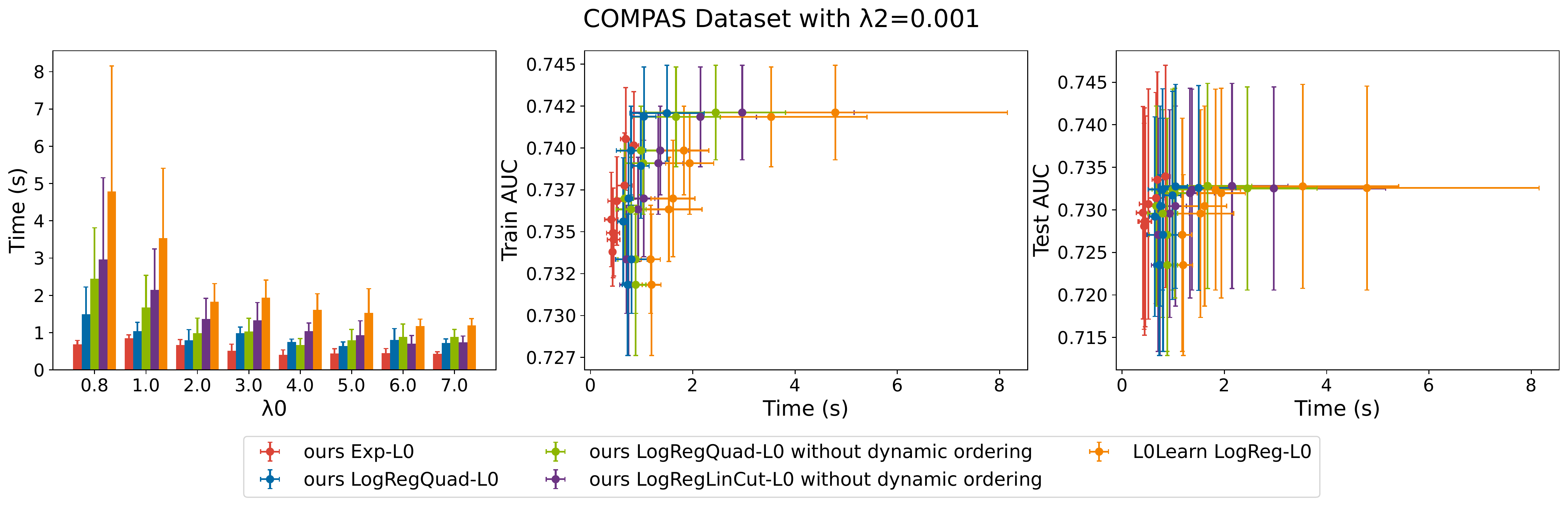}

  \caption{Computational times of different methods. ``Exp'' stands for exponential loss, ``LogReg'' stands for logistic loss, and ``Quad'' stands for quadratic cuts. Note that there is no $\ell_2$ penalty for the exponential loss. \textit{Our Exp-L0 method is generally about 4 times faster than L0Learn.} Note that the AUC axes indicate practically similar performance for these particular methods; the training time is what differentiates the methods.}
  \label{fig:time_speed_up_compas}
\end{figure*}

\begin{figure*}[t]
  \centering
  \includegraphics[width=1.0\linewidth]{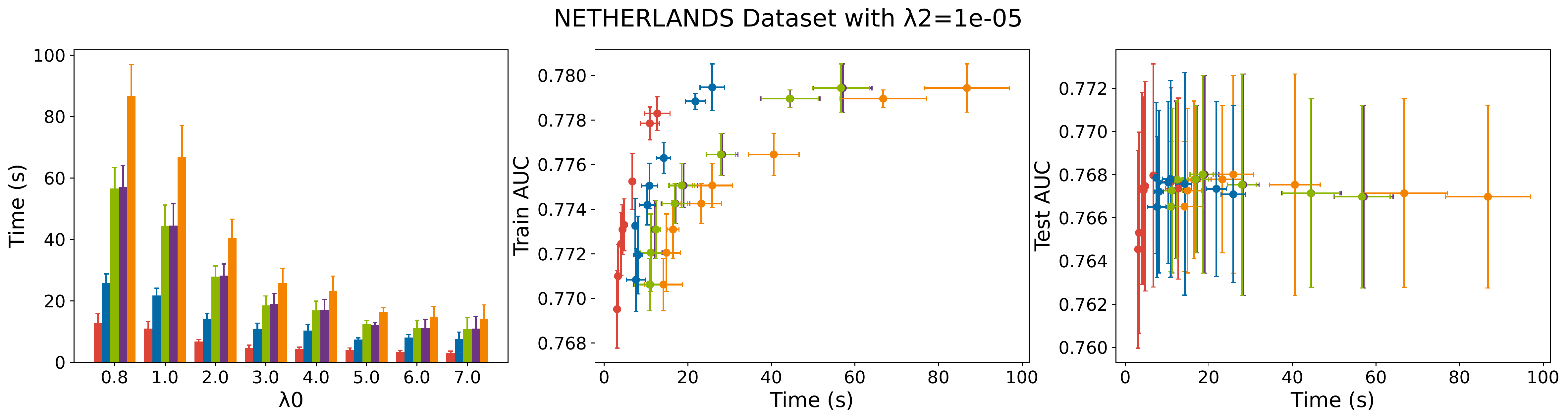}

  \includegraphics[width=1.0\linewidth]{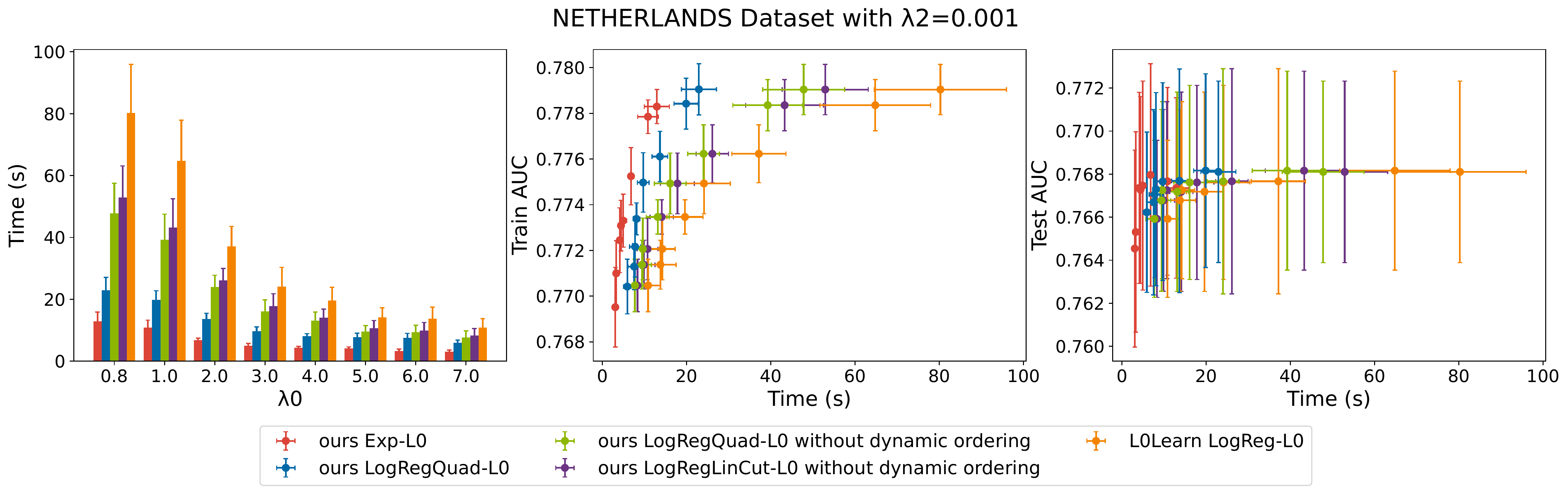}

  \caption{Computational times of different methods. ``Exp'' stands for exponential loss, ``LogReg'' stands for logistic loss, and ``Quad'' stands for quadratic cuts. Note that there is no $\ell_2$ penalty for the exponential loss. \textit{Our Exp-L0 method is generally about 4 times faster than L0Learn.} Note that the AUC axes indicate practically similar performance for these particular methods; the training time is what differentiates the methods.}
  \label{fig:time_speed_up_netherlands}
\end{figure*}

We show results on time savings from our first method (linear cut, quadratic cut, and dynamic ordering) and our second method (exponential loss). The general trends are:
\emph{\romannumeral1}) Using the quadratic cut makes the algorithm faster than the linear cut, i.e., there is more time saved with stronger $\ell_2$ regularization.
\emph{\romannumeral2}) Using dynamic ordering and the quadratic cut together makes the algorithm much faster than using the quadratic cut alone.
\emph{\romannumeral3}) When features are binary, using the exponential loss has the greatest computational advantage.
These trends are shown fairly uniformly across datasets. Results for each dataset are shown in Figures \ref{fig:time_speed_up_compas}-\ref{fig:time_speed_up_netherlands}.

\subsection{Support versus AUC, Accuracy, and F1 Score}\label{app:support_tradeoffs}

\begin{figure*}[ht]
  \centering
    \includegraphics[width=1.0\textwidth]{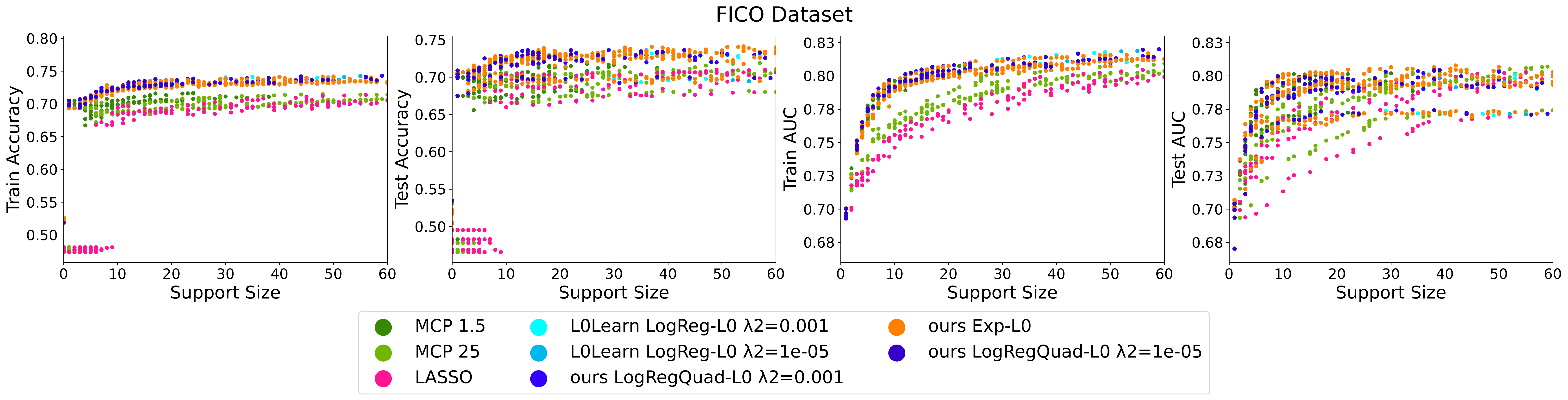}
\caption{Results on the FICO dataset. See Figure 5 in the main paper for results on the COMPAS and NETHERLANDS datasets. The L0Learn points are mostly overlapping with points from our methods.}
\label{fig:real_data_neterlands}
\end{figure*}

We provide the full regularization paths for the FICO dataset (see Figure \ref{fig:real_data_neterlands}). Our first method (quadratic cut + dynamic ordering) and second method (exponential loss) obtain high-quality solutions and their AUC and accuracy curves are similar to those from other methods. Our methods have computational advantages over L0Learn, as shown in Section \ref{app:run_time_savings}.

\begin{figure*}[ht]
    \centering
    \includegraphics[width=1.0\textwidth]{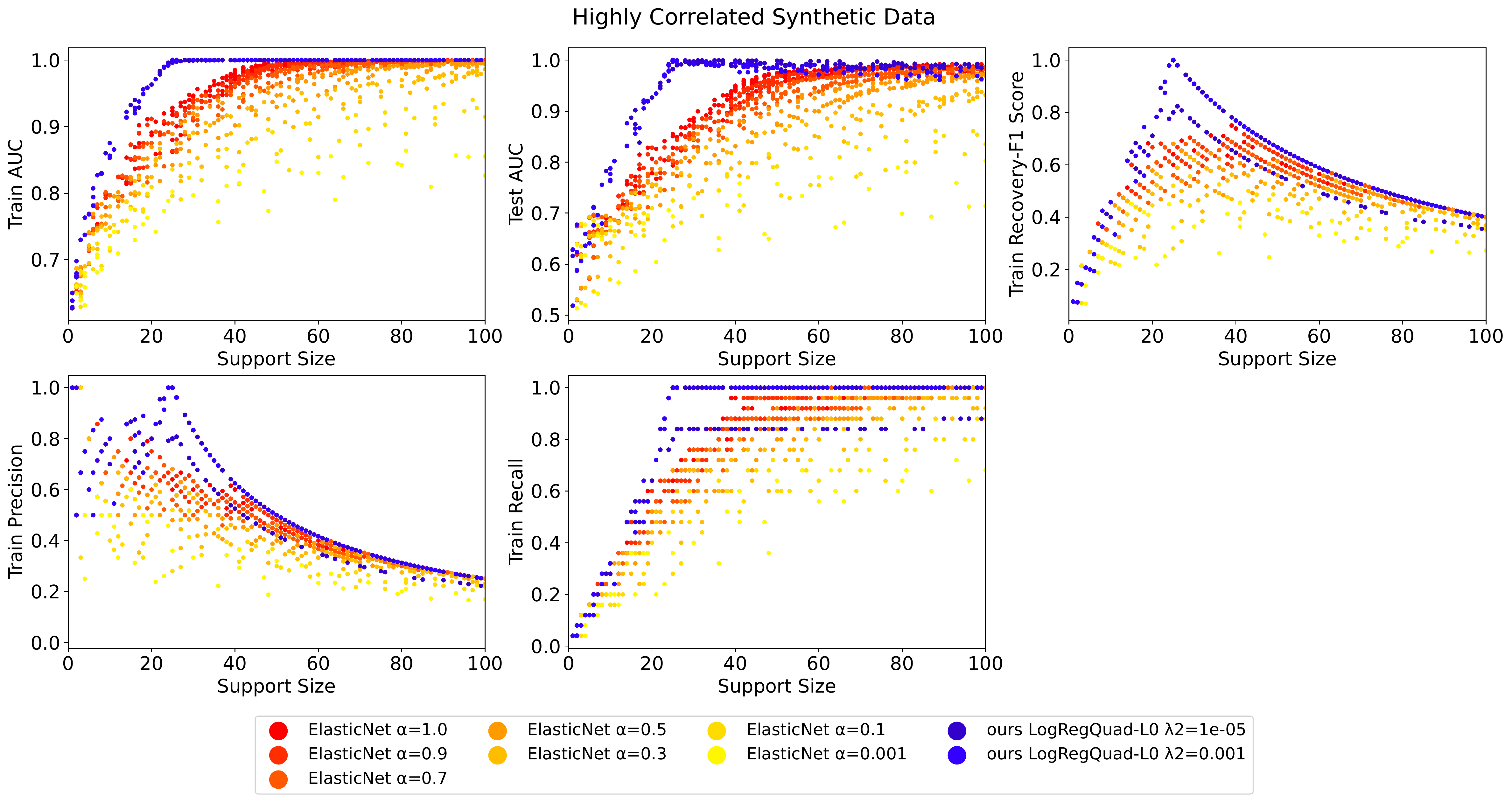}

\caption{ElasticNet and our methods' results on the highly correlated synthetic dataset. }
\label{fig:synthetic_data_elastic}
\end{figure*}

To investigate whether a small $\ell_2$ regularization would help with the LASSO baseline, we provide a comparison between our method and the the ElasticNet method on the highly correlated synthetic dataset for the classification task. We used the glmnet R package for the ElasticNet baseline. The hyperparameter $\alpha \in [0, 1]$ controls the balance between $\ell_1$ and $\ell_2$ regularization.  The LASSO method corresponds to $\alpha=1.0$. Besides $\alpha=1.0$, we also consider $\alpha \in \{0.9, 0.7, 0.5, 0.3, 0.1, 0.001\}$. As shown in Figure~\ref{fig:synthetic_data_elastic}, a small $\ell_2$ regularization term does not improve the LASSO method much. When $\alpha$ decreases, the solution quality degrades. This is potentially because more $\ell_2$ regularization leads to non-sparse solutions with small coefficients, neither of which will lead to better performance here. 

\subsection{Samples of Sparse Models on the FICO and NETHERLANDS datasets}

We provide some sample sparse models produced by minimizing the exponential loss and minimizing the logistic loss (quadratic cut + dynamic ordering) on the FICO and NETHERLANDS datasets.

The FICO dataset has 10459 samples and 1917 features. The NETHERLANDS dataset has 20000 samples and 2024 features. All models were developed from the third fold of our 5-fold cross validation split.

\begin{figure}
\textbf{FICO model using the exponential loss:}\\

$\lambda_0=5$:
\begin{align*}
    score = & -0.2584626 \\
&+0.1825955 \times \bm{1}_{ A  \leq 63 } +0.1387806 \times \bm{1}_{ A  \leq 70 } \\
&+0.2286364 \times \bm{1}_{ A  \leq 74 } +0.2569742 \times \bm{1}_{ A  \leq 83 }  && \textit{\# A :ExternalRiskEstimate } \\
&+0.1840013 \times \bm{1}_{ B  \leq 51 } +0.172138 \times \bm{1}_{ B  \leq 75 }  && \textit{\# B :AverageMInFile } \\
&+0.2015039 \times \bm{1}_{ C  \leq 13 } +0.1923697 \times \bm{1}_{ C  \leq 31 }  && \textit{\# C :NumSatisfactoryTrades } \\
&+0.2654667 \times \bm{1}_{ D  \leq 96 }  && \textit{\# D :PercentTradesNeverDelq } \\
&+0.2320259 \times \bm{1}_{ E  \leq 33 }  && \textit{\# E :MSinceMostRecentDelq } \\
&+0.1009372 \times \bm{1}_{ F  \leq 8 }  && \textit{\# F :NumTotalTrades } \\
&-0.2311165 \times \bm{1}_{ G  \leq 46 }  && \textit{\# G :PercentInstallTrades } \\
&-0.7723769 \times \bm{1}_{ H  \leq -8 } +0.3636577 \times \bm{1}_{ H  \leq 0 }  && \textit{\# H :MSinceMostRecentInqexcl7days } \\
&-0.2762694 \times \bm{1}_{ I  \leq 5 }  && \textit{\# I :NumInqLast6M } \\
&-0.1897788 \times \bm{1}_{ J  \leq 37 } -0.2742168 \times \bm{1}_{ J  \leq 73 }  && \textit{\# J :NetFractionRevolvingBurden } \\
&-0.1038025 \times \bm{1}_{ K  \leq 5 } -0.1938047 \times \bm{1}_{ K  \leq 7 }  && \textit{\# K :NumRevolvingTradesWBalance }
\end{align*}
    \centering
    \includegraphics[width=1.0\textwidth]{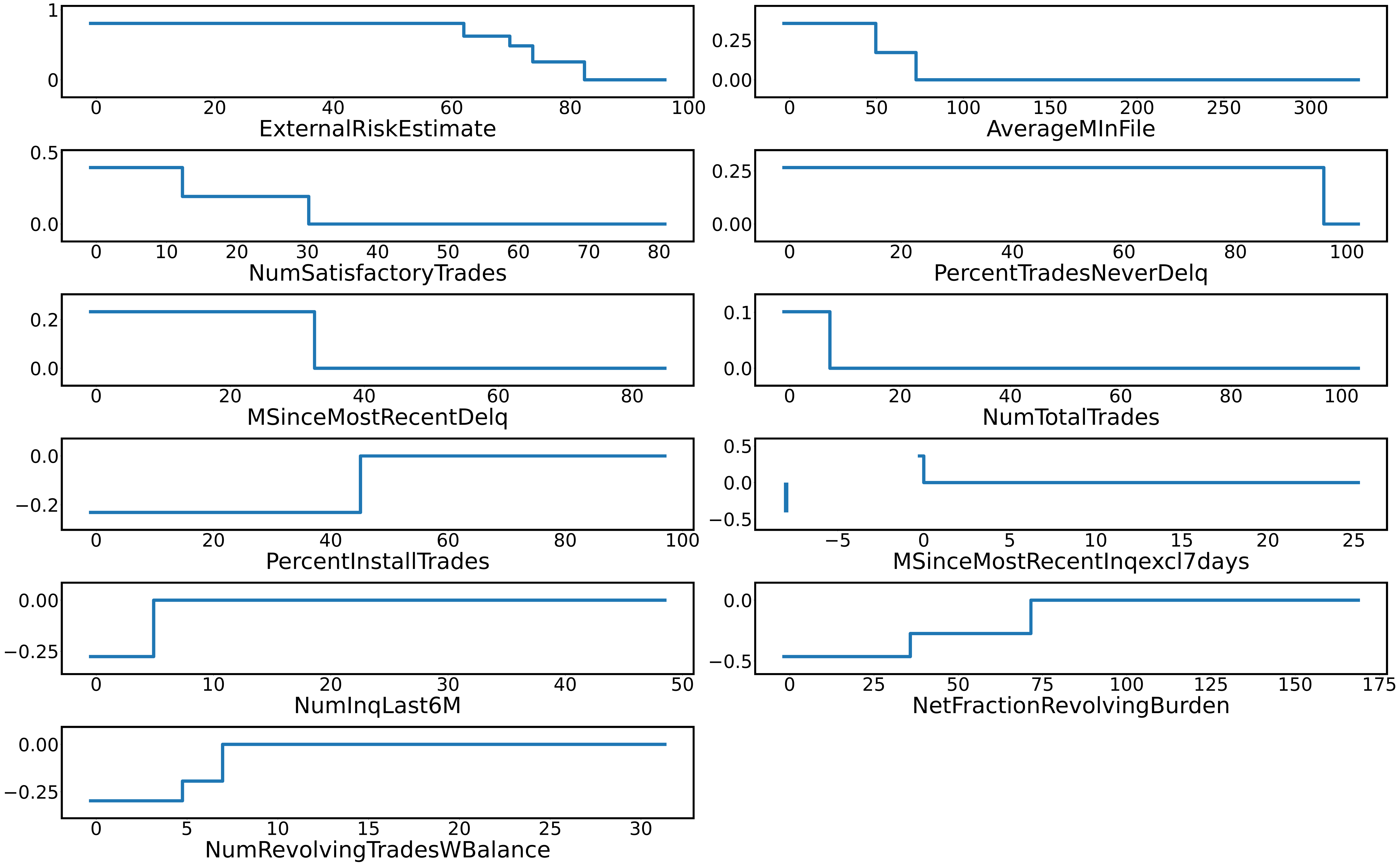}
    \caption{FICO score contributions with the exponential loss and $\lambda_0=5$. Training duration is 3.15 seconds. Note that no monotonicity constraints were imposed.}
    \label{fig:sparse_model_fico_expo}
\end{figure}

\begin{figure}[ht]
\textbf{FICO model using the logistic loss (quadratic cut + dynamic ordering):}

$\lambda_0=5$, $\lambda_2=0.001$:
\begin{align*}
    score = & 2.805021 \\
&+0.4071199 \times \bm{1}_{ A  \leq 63 } +0.310368 \times \bm{1}_{ A  \leq 70 } \\ &+0.4604512 \times \bm{1}_{ A  \leq 74 } +0.5471219 \times \bm{1}_{ A  \leq 83 }  && \textit{\# A :ExternalRiskEstimate } \\
&+0.408959 \times \bm{1}_{ B  \leq 51 } +0.3283239 \times \bm{1}_{ B  \leq 75 }  && \textit{\# B :AverageMInFile } \\
&+0.4225237 \times \bm{1}_{ C  \leq 13 } +0.3396898 \times \bm{1}_{ C  \leq 31 }  && \textit{\# C :NumSatisfactoryTrades } \\
&+0.525166 \times \bm{1}_{ D  \leq 96 }  && \textit{\# D :PercentTradesNeverDelq } \\
&+0.4427697 \times \bm{1}_{ E  \leq 33 }  && \textit{\# E :MSinceMostRecentDelq } \\
&-0.4317725 \times \bm{1}_{ F  \leq 46 }  && \textit{\# F :PercentInstallTrades } \\
&-1.576435 \times \bm{1}_{ G  \leq -8 } +0.5045199 \times \bm{1}_{ G  \leq 0 } \\
&+0.2874494 \times \bm{1}_{ G  \leq 1 }  && \textit{\# G :MSinceMostRecentInqexcl7days } \\
&-3.97116 \times \bm{1}_{ H  \leq 11 }  && \textit{\# H :NumInqLast6M } \\
&-0.3657186 \times \bm{1}_{ I  \leq 37 } -0.5681891 \times \bm{1}_{ I  \leq 73 }  && \textit{\# I :NetFractionRevolvingBurden } \\
&-0.4969551 \times \bm{1}_{ J  \leq 7 }  && \textit{\# J :NumRevolvingTradesWBalance }
\end{align*}
    \centering
    \includegraphics[width=1.0\textwidth]{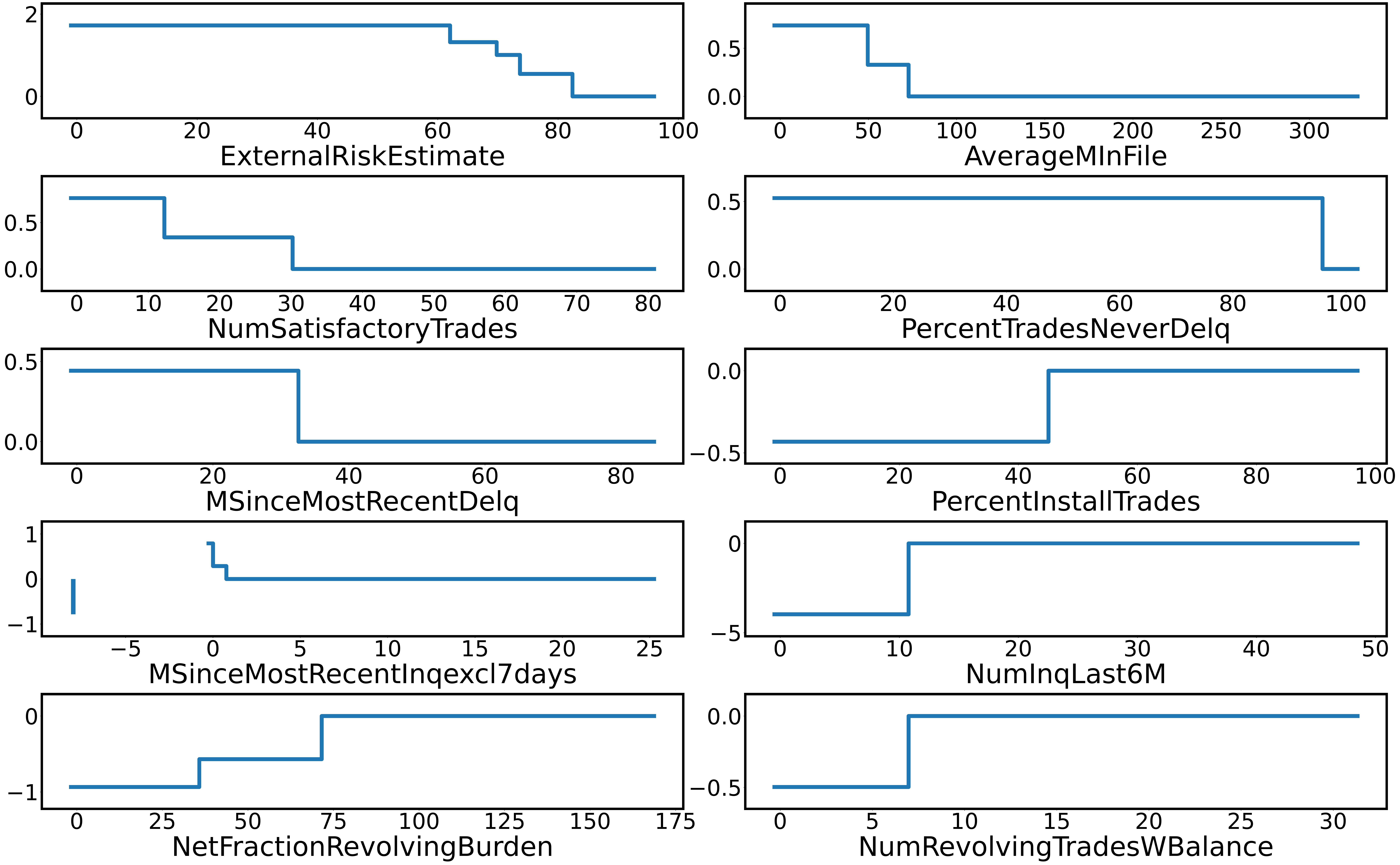}
    \caption{FICO score contributions with the logistic loss and $\lambda_0=5, \lambda_2=0.001$. Training duration is 5.95 seconds.}
    \label{fig:sparse_model_fico_logistic}
\end{figure}

\textbf{FICO Baseline Performance:} The sparse models below approximately match the performance of black-box models shown in previous works~\citep{ChenEtAl21}. We also ran a GBDT model~\citep{friedman2001greedy} with max depth set to be 3 and number of boosting stages set to be 100. The AUC on the training set is $0.8318 \pm 0.0028$, and the AUC on the test set is $0.7959 \pm 0.0133$. The result on the test set is comparable to what we have shown in Figure~\ref{fig:time_speed_up}. The models are in Figures \ref{fig:sparse_model_fico_expo}-\ref{fig:sparse_model_fico_logistic}.

\begin{figure}[ht]
\textbf{NETHERLANDS model using the exponential loss:}\\

$\lambda_0=7$:
\begin{align*}
    score = & 3.114342 \\
&-0.1439394 \times \bm{1}_{ A  == female }  && \textit{\# A :sex } \\
&-0.4739628 \times \bm{1}_{ B  \leq 0.0 } -0.3336059 \times \bm{1}_{ B  \leq 1.098612289 } \\
&-0.3083761 \times \bm{1}_{ B  \leq 1.609437912 }  && \textit{\# B :log \# of previous penal cases } \\
&+0.2887266 \times \bm{1}_{ C  \leq 22.26146475 } +0.2354507 \times \bm{1}_{ C  \leq 28.48266213076 } \\
&+0.1951787 \times \bm{1}_{ C  \leq 39.07432555374 } +0.2304243 \times \bm{1}_{ C  \leq 46.91581109 }  && \textit{\# C :age in years } \\
&-0.1674992 \times \bm{1}_{ D  \leq 28.069209540720003 }  && \textit{\# D :age at first penal case } \\
&+0.1526613 \times \bm{1}_{ E  \leq 7.0 }  && \textit{\# E :offence type } \\
&-1.381801 \times \bm{1}_{ F  \leq 0.0 }  && \textit{\# F :11-20 previous case } \\
&-1.856642 \times \bm{1}_{ G  \leq 0.0 }  && \textit{\# G :$>$20 previous case }
\end{align*}
\centering
    \includegraphics[width=1.0\textwidth]{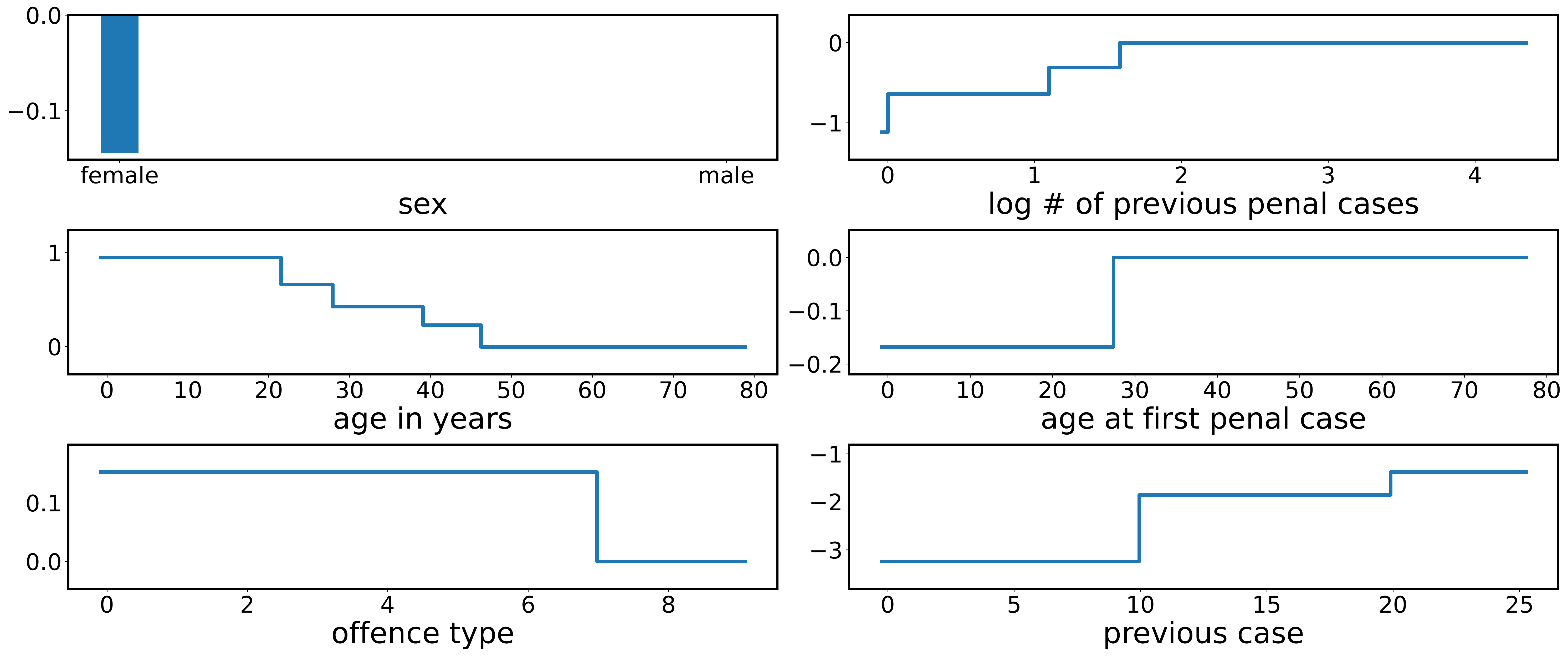}
    \caption{NETHERLANDS score contributions with the exponential loss and $\lambda_0=7$. Training duration is 2.73 seconds. Note that there are no monotonicity constraints imposed.}
    \label{fig:sparse_model_netherlands_expo}
\end{figure}

\begin{figure}[ht]
\textbf{NETHERLANDS model using the logistic loss (quadratic cut + dynamic ordering):}

$\lambda_0=7$, $\lambda_2=0.001$:
\begin{align*}
    score = & 6.259994 \\
&-0.3092142 \times \bm{1}_{ A  == female }  && \textit{\# A :sex } \\
&-0.7374188 \times \bm{1}_{ B  \leq 0.0 } -0.4302188 \times \bm{1}_{ B  \leq 0.693147181 } \\
&-0.2888496 \times \bm{1}_{ B  \leq 1.098612289 } -0.3933033 \times \bm{1}_{ B  \leq 1.386294361 } \\
&-0.5383587 \times \bm{1}_{ B  \leq 1.945910149 }  && \textit{\# B :log \# of previous penal cases } \\
&+0.3877289 \times \bm{1}_{ C  \leq 18.94046991832 } +0.5554352 \times \bm{1}_{ C  \leq 23.01017483608 } \\
&+0.4700141 \times \bm{1}_{ C  \leq 31.552317465359998 } +0.6324188 \times \bm{1}_{ C  \leq 43.91512663 }  && \textit{\# C :age in years } \\
&-0.2645467 \times \bm{1}_{ D  \leq 27.986380572549994 }  && \textit{\# D :age at first penal case } \\
&+0.2861914 \times \bm{1}_{ E  \leq 7.0 }  && \textit{\# E :offence type } \\
&-2.655844 \times \bm{1}_{ F  \leq 0.0 }  && \textit{\# F :11-20 previous case } \\
&-3.605789 \times \bm{1}_{ G  \leq 0.0 }  && \textit{\# G :$>$20 previous case }
\end{align*}
\centering
\includegraphics[width=1.0\textwidth]{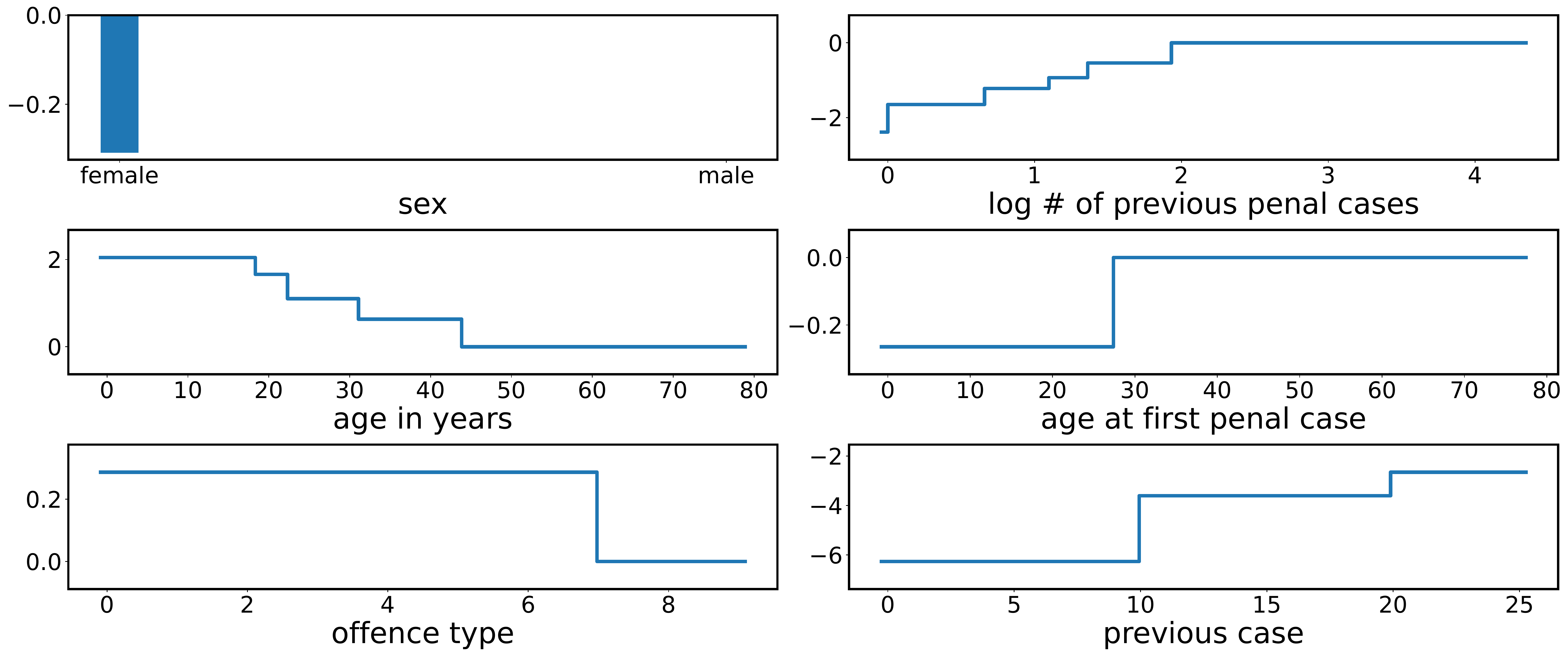}
\caption{NETHERLANDS model score contributions with logistic loss, $\lambda_0=7$, and $\lambda_2=0.001$. Training duration is 7.2 seconds.}
\label{fig:sparse_model_netherlands_logistic}
\end{figure}

\textbf{NETHERLANDS Baseline Performance:}
 The sparse models below approximately match the performance of black-box models. We ran a GBDT model~\citep{friedman2001greedy} with max depth set to be 3 and number of boosting stages set to be 100. The AUC on the training set is $0.7850 \pm 0.0014$, and the AUC on the test set is $0.7696 \pm 0.0062$. The result on the test set is comparable to what we have shown in Figure~\ref{fig:time_speed_up_netherlands}. (For the NETHERLANDS dataset, ages are collected in terms of months. That is why age thresholds are shown with float numbers.) The models are in Figures \ref{fig:sparse_model_netherlands_expo}-\ref{fig:sparse_model_netherlands_logistic}.

%% file: main.bbl
\begin{thebibliography}{45}
\providecommand{\natexlab}[1]{#1}
\providecommand{\url}[1]{\texttt{#1}}
\expandafter\ifx\csname urlstyle\endcsname\relax
  \providecommand{\doi}[1]{doi: #1}\else
  \providecommand{\doi}{doi: \begingroup \urlstyle{rm}\Url}\fi

\bibitem[Bahmani et~al.(2013)Bahmani, Raj, and Boufounos]{bahmani2013greedy}
Sohail Bahmani, Bhiksha Raj, and Petros~T Boufounos.
\newblock Greedy sparsity-constrained optimization.
\newblock \emph{Journal of Machine Learning Research}, 14\penalty0
  (Mar):\penalty0 807--841, 2013.

\bibitem[Beck and Eldar(2013)]{beck2013sparsity}
Amir Beck and Yonina~C Eldar.
\newblock Sparsity constrained nonlinear optimization: Optimality conditions
  and algorithms.
\newblock \emph{SIAM Journal on Optimization}, 23\penalty0 (3):\penalty0
  1480--1509, 2013.

\bibitem[Bertsekas(1997)]{bertsekas1997nonlinear}
Dimitri~P Bertsekas.
\newblock Nonlinear programming.
\newblock \emph{Journal of the Operational Research Society}, 48\penalty0
  (3):\penalty0 334--334, 1997.

\bibitem[Bertsimas and King(2017)]{bertsimas2017logistic}
Dimitris Bertsimas and Angela King.
\newblock Logistic regression: From art to science.
\newblock \emph{Statistical Science}, pages 367--384, 2017.

\bibitem[Bertsimas et~al.(2021)Bertsimas, Pauphilet, and
  Van~Parys]{bertsimas2017sparse}
Dimitris Bertsimas, Jean Pauphilet, and Bart Van~Parys.
\newblock Sparse classification: a scalable discrete optimization perspective.
\newblock \emph{Machine Learning}, 110\penalty0 (11):\penalty0 3177--3209,
  2021.

\bibitem[Blumensath and Davies(2009)]{blumensath2009iterative}
Thomas Blumensath and Mike~E Davies.
\newblock Iterative hard thresholding for compressed sensing.
\newblock \emph{Applied and Computational Harmonic Analysis}, 27\penalty0
  (3):\penalty0 265--274, 2009.

\bibitem[Breheny and Huang(2011)]{ncvreg}
Patrick Breheny and Jian Huang.
\newblock Coordinate descent algorithms for nonconvex penalized regression,
  with applications to biological feature selection.
\newblock \emph{Annals of Applied Statistics}, 5\penalty0 (1):\penalty0
  232--253, 2011.

\bibitem[Chen et~al.(2021)Chen, Lin, Rudin, Shaposhnik, Wang, and
  Wang]{ChenEtAl21}
Chaofan Chen, Kangcheng Lin, Cynthia Rudin, Yaron Shaposhnik, Sijia Wang, and
  Tong Wang.
\newblock A holistic approach to interpretability in financial lending: Models,
  visualizations, and summary-explanations.
\newblock \emph{Decision Support Systems}, page 113647, 2021.

\bibitem[Daubechies et~al.(2004)Daubechies, Defrise, and
  De~Mol]{daubechies2004iterative}
Ingrid Daubechies, Michel Defrise, and Christine De~Mol.
\newblock An iterative thresholding algorithm for linear inverse problems with
  a sparsity constraint.
\newblock \emph{Communications on Pure and Applied Mathematics: A Journal
  Issued by the Courant Institute of Mathematical Sciences}, 57\penalty0
  (11):\penalty0 1413--1457, 2004.

\bibitem[Dedieu et~al.(2021)Dedieu, Hazimeh, and Mazumder]{dedieu2020learning}
Antoine Dedieu, Hussein Hazimeh, and Rahul Mazumder.
\newblock Learning sparse classifiers: Continuous and mixed integer
  optimization perspectives.
\newblock \emph{Journal of Machine Learning Research}, 22\penalty0
  (135):\penalty0 1--47, 2021.

\bibitem[Del~Moral et~al.(2006)Del~Moral, Doucet, and Jasra]{del2006sequential}
Pierre Del~Moral, Arnaud Doucet, and Ajay Jasra.
\newblock Sequential monte carlo samplers.
\newblock \emph{Journal of the Royal Statistical Society: Series B (Statistical
  Methodology)}, 68\penalty0 (3):\penalty0 411--436, 2006.

\bibitem[Dempster et~al.(1977)Dempster, Laird, and Rubin]{dempster1977maximum}
Arthur~P Dempster, Nan~M Laird, and Donald~B Rubin.
\newblock Maximum likelihood from incomplete data via the em algorithm.
\newblock \emph{Journal of the Royal Statistical Society: Series B
  (Methodological)}, 39\penalty0 (1):\penalty0 1--22, 1977.

\bibitem[Elenberg et~al.(2018)Elenberg, Khanna, Dimakis, and
  Negahban]{elenberg2018restricted}
Ethan~R Elenberg, Rajiv Khanna, Alexandros~G Dimakis, and Sahand Negahban.
\newblock Restricted strong convexity implies weak submodularity.
\newblock \emph{The Annals of Statistics}, 46\penalty0 (6B):\penalty0
  3539--3568, 2018.

\bibitem[Ertekin and Rudin(2011)]{ErtekinRu11}
{\c{S}eyda}~Ertekin and Cynthia Rudin.
\newblock On equivalence relationships between classification and ranking
  algorithms.
\newblock \emph{Journal of Machine Learning Research}, 12:\penalty0 2905--2929,
  2011.

\bibitem[{FICO} et~al.(2018){FICO}, {Google}, {Imperial College London}, {MIT},
  {University of Oxford}, {UC Irvine}, and {UC Berkeley}]{fico}
{FICO}, {Google}, {Imperial College London}, {MIT}, {University of Oxford}, {UC
  Irvine}, and {UC Berkeley}.
\newblock {Explainable Machine Learning Challenge}.
\newblock
  \url{https://community.fico.com/s/explainable-machine-learning-challenge},
  2018.

\bibitem[Freund and Schapire(1997)]{freund1997decision}
Yoav Freund and Robert~E Schapire.
\newblock A decision-theoretic generalization of on-line learning and an
  application to boosting.
\newblock \emph{Journal of Computer and System Sciences}, 55\penalty0
  (1):\penalty0 119--139, 1997.

\bibitem[Friedman et~al.(2010)Friedman, Hastie, and Tibshirani]{glmnet}
Jerome Friedman, Trevor Hastie, and Robert Tibshirani.
\newblock Regularization paths for generalized linear models via coordinate
  descent.
\newblock \emph{Journal of Statistical Software}, 33\penalty0 (1):\penalty0
  1--22, 2010.

\bibitem[Friedman(2001)]{friedman2001greedy}
Jerome~H Friedman.
\newblock Greedy function approximation: a gradient boosting machine.
\newblock \emph{Annals of Statistics}, pages 1189--1232, 2001.

\bibitem[Gilmore and Gomory(1961)]{gilmore1961linear}
Paul~C Gilmore and Ralph~E Gomory.
\newblock A linear programming approach to the cutting-stock problem.
\newblock \emph{Operations Research}, 9\penalty0 (6):\penalty0 849--859, 1961.

\bibitem[Gilmore and Gomory(1963)]{gilmore1963linear}
Paul~C Gilmore and Ralph~E Gomory.
\newblock A linear programming approach to the cutting stock problem—part ii.
\newblock \emph{Operations Research}, 11\penalty0 (6):\penalty0 863--888, 1963.

\bibitem[Hastie and Tibshirani(2017)]{hastie2017generalized}
Trevor~J Hastie and Robert~J Tibshirani.
\newblock \emph{Generalized additive models}.
\newblock Routledge, 2017.

\bibitem[Hazimeh and Mazumder(2020)]{hazimeh2020fast}
Hussein Hazimeh and Rahul Mazumder.
\newblock Fast best subset selection: Coordinate descent and local
  combinatorial optimization algorithms.
\newblock \emph{Operations Research}, 68\penalty0 (5):\penalty0 1517--1537,
  2020.

\bibitem[Kelley(1960)]{kelley1960cutting}
James~E Kelley, Jr.
\newblock The cutting-plane method for solving convex programs.
\newblock \emph{Journal of the Society for Industrial and Applied Mathematics},
  8\penalty0 (4):\penalty0 703--712, 1960.

\bibitem[Kirkpatrick et~al.(1983)Kirkpatrick, Gelatt, and
  Vecchi]{kirkpatrick1983optimization}
Scott Kirkpatrick, C~Daniel Gelatt, and Mario~P Vecchi.
\newblock Optimization by simulated annealing.
\newblock \emph{Science}, 220\penalty0 (4598):\penalty0 671--680, 1983.

\bibitem[Land and Doig(2010)]{land2010automatic}
Ailsa~H Land and Alison~G Doig.
\newblock An automatic method for solving discrete programming problems.
\newblock In \emph{50 Years of Integer Programming 1958-2008}, pages 105--132.
  Springer, 2010.

\bibitem[Larson et~al.(2016)Larson, Mattu, Kirchner, and
  Angwin]{LarsonMaKiAn16}
J.~Larson, S.~Mattu, L.~Kirchner, and J.~Angwin.
\newblock How we analyzed the {COMPAS} recidivism algorithm.
\newblock \emph{ProPublica}, 2016.

\bibitem[Lou et~al.(2016)Lou, Bien, Caruana, and Gehrke]{lou2016sparse}
Yin Lou, Jacob Bien, Rich Caruana, and Johannes Gehrke.
\newblock Sparse partially linear additive models.
\newblock \emph{Journal of Computational and Graphical Statistics}, 25\penalty0
  (4):\penalty0 1126--1140, 2016.

\bibitem[Lozano et~al.(2011)Lozano, Swirszcz, and Abe]{lozano2011group}
Aurelie Lozano, Grzegorz Swirszcz, and Naoki Abe.
\newblock Group orthogonal matching pursuit for logistic regression.
\newblock In \emph{Proceedings of the Fourteenth International Conference on
  Artificial Intelligence and Statistics}, pages 452--460, 2011.

\bibitem[Metropolis et~al.(1953)Metropolis, Rosenbluth, Rosenbluth, Teller, and
  Teller]{metropolis1953equation}
Nicholas Metropolis, Arianna~W Rosenbluth, Marshall~N Rosenbluth, Augusta~H
  Teller, and Edward Teller.
\newblock Equation of state calculations by fast computing machines.
\newblock \emph{The Journal of Chemical Physics}, 21\penalty0 (6):\penalty0
  1087--1092, 1953.

\bibitem[Nori et~al.(2019)Nori, Jenkins, Koch, and
  Caruana]{nori2019interpretml}
Harsha Nori, Samuel Jenkins, Paul Koch, and Rich Caruana.
\newblock Interpretml: A unified framework for machine learning
  interpretability.
\newblock \emph{arXiv preprint arXiv:1909.09223}, 2019.

\bibitem[Patrascu and Necoara(2015)]{patrascu2015random}
Andrei Patrascu and Ion Necoara.
\newblock Random coordinate descent methods for $\ell_ {0}$ regularized convex
  optimization.
\newblock \emph{IEEE Transactions on Automatic Control}, 60\penalty0
  (7):\penalty0 1811--1824, 2015.

\bibitem[Platt(1998)]{platt1998sequential}
John Platt.
\newblock Sequential minimal optimization: A fast algorithm for training
  support vector machines.
\newblock Technical Report MSR-TR-98-14, April 21 1998.

\bibitem[Rudin et~al.(2022)Rudin, Chen, Chen, Huang, Semenova, and
  Zhong]{rudin2022interpretable}
Cynthia Rudin, Chaofan Chen, Zhi Chen, Haiyang Huang, Lesia Semenova, and Chudi
  Zhong.
\newblock Interpretable machine learning: Fundamental principles and 10 grand
  challenges.
\newblock \emph{Statistics Surveys}, 16:\penalty0 1--85, 2022.

\bibitem[Sakaue and Marumo(2019)]{sakaue2019best}
Shinsaku Sakaue and Naoki Marumo.
\newblock Best-first search algorithm for non-convex sparse minimization.
\newblock \emph{arXiv preprint arXiv:1910.01296}, 2019.

\bibitem[Sato et~al.(2016)Sato, Takano, Miyashiro, and
  Yoshise]{sato2016feature}
Toshiki Sato, Yuichi Takano, Ryuhei Miyashiro, and Akiko Yoshise.
\newblock Feature subset selection for logistic regression via mixed integer
  optimization.
\newblock \emph{Computational Optimization and Applications}, 64\penalty0
  (3):\penalty0 865--880, 2016.

\bibitem[Sato et~al.(2017)Sato, Takano, and Miyashiro]{sato2017piecewise}
Toshiki Sato, Yuichi Takano, and Ryuhei Miyashiro.
\newblock Piecewise-linear approximation for feature subset selection in a
  sequential logit model.
\newblock \emph{Journal of the Operations Research Society of Japan},
  60\penalty0 (1):\penalty0 1--14, 2017.

\bibitem[Schapire and Freund(2013)]{schapire2013boosting}
Robert~E Schapire and Yoav Freund.
\newblock Boosting: Foundations and algorithms.
\newblock \emph{Kybernetes}, 2013.

\bibitem[Tibshirani(1996)]{tibshirani1996regression}
Robert Tibshirani.
\newblock Regression shrinkage and selection via the lasso.
\newblock \emph{Journal of the Royal Statistical Society: Series B
  (Methodological)}, 58\penalty0 (1):\penalty0 267--288, 1996.

\bibitem[Tollenaar and Van~der Heijden(2013)]{tollenaar2013method}
Nikolaj Tollenaar and PGM Van~der Heijden.
\newblock Which method predicts recidivism best?: a comparison of statistical,
  machine learning and data mining predictive models.
\newblock \emph{Journal of the Royal Statistical Society: Series A (Statistics
  in Society)}, 176\penalty0 (2):\penalty0 565--584, 2013.

\bibitem[Ustun and Rudin(2017)]{ustun2017optimized}
Berk Ustun and Cynthia Rudin.
\newblock Optimized risk scores.
\newblock In \emph{Proceedings of the 23rd {ACM} {SIGKDD} International
  Conference on Knowledge Discovery and Data Mining}, pages 1125--1134, 2017.

\bibitem[Ustun and Rudin(2019)]{ustun2019learning}
Berk Ustun and Cynthia Rudin.
\newblock Learning optimized risk scores.
\newblock \emph{J. Mach. Learn. Res.}, 20:\penalty0 150--1, 2019.

\bibitem[Wolsey and Nemhauser(1999)]{wolsey1999integer}
Laurence~A Wolsey and George~L Nemhauser.
\newblock \emph{Integer and Combinatorial Optimization}, volume~55.
\newblock John Wiley \& Sons, 1999.

\bibitem[Zhang et~al.(2021)Zhang, Zhu, Zhu, and Wang]{zhang2021certifiably}
Yanhang Zhang, Junxian Zhu, Jin Zhu, and Xueqin Wang.
\newblock Certifiably polynomial algorithm for best group subset selection.
\newblock \emph{arXiv preprint arXiv:2104.12576}, 2021.
\newblock Code version: December 8, 2021.

\bibitem[Zhou et~al.(2021)Zhou, Xiu, and Qi]{zhou2021global}
Shenglong Zhou, Naihua Xiu, and Hou-Duo Qi.
\newblock Global and quadratic convergence of newton hard-thresholding pursuit.
\newblock \emph{J. Mach. Learn. Res.}, 22\penalty0 (12):\penalty0 1--45, 2021.

\bibitem[Zhu et~al.(2020)Zhu, Wen, Zhu, Zhang, and Wang]{zhu2020polynomial}
Junxian Zhu, Canhong Wen, Jin Zhu, Heping Zhang, and Xueqin Wang.
\newblock A polynomial algorithm for best-subset selection problem.
\newblock \emph{Proceedings of the National Academy of Sciences}, 117\penalty0
  (52):\penalty0 33117--33123, 2020.

\end{thebibliography}
